\documentclass[letterpaper, conference]{IEEEtran}
\IEEEoverridecommandlockouts
\def\BibTeX{{\rm B\kern-.05em{\sc i\kern-.025em b}\kern-.08em
    T\kern-.1667em\lower.7ex\hbox{E}\kern-.125emX}}

\makeatletter
\def\ps@headings{%
\def\@oddhead{\mbox{}\scriptsize\rightmark \hfil \thepage}%
\def\@evenhead{\scriptsize\thepage \hfil \leftmark\mbox{}}%
\def\@oddfoot{}%
\def\@evenfoot{}}
\makeatother
\usepackage{graphicx}
\usepackage{cite}
\usepackage{amsmath,amssymb,amsfonts}
\usepackage{textcomp}
\usepackage{xcolor}
\usepackage{comment}

\usepackage[caption=false]{subfig}
\usepackage[ruled,vlined]{algorithm2e}
\usepackage{bbm}

\usepackage[
 top=0.75in,
 left=0.630in,
 right=0.640in,
 bottom=1.06in,
 ]{geometry}

\usepackage[strings]{underscore}

\newcommand{\revision}[1]{\textcolor{black}{#1}}

\newcommand{\myparagraph}[1]{ \noindent \textbf{#1.}}

\newif\iffullpaper
\fullpaperfalse

\begin{document}
\title{Backdoor Attacks in Peer-to-Peer Federated Learning}

\author{%
  \IEEEauthorblockN{%
    Georgios Syros$^{*}$\thanks{$^{*}$ These authors contributed equally.},
    Gökberk Yar$^{*}$,
    Simona Boboila,
    Cristina Nita-Rotaru,
    Alina Oprea
  }%
  \IEEEauthorblockA{Northeastern University, Khoury College of Computer Sciences}%
}

\maketitle


\begin{abstract}
Most machine learning applications rely on centralized learning processes, opening up the risk of exposure of their training datasets. While federated learning (FL) mitigates to some extent these privacy risks, it relies on a trusted aggregation server for training a shared global model. Recently, new distributed learning architectures based on Peer-to-Peer Federated Learning (P2PFL) offer advantages in terms of both privacy and reliability. Still, their resilience to poisoning attacks during training has not been investigated. In this paper, we propose new backdoor attacks for P2PFL that leverage structural graph properties to select the malicious nodes, and achieve high attack success, while remaining stealthy. We evaluate our attacks under various realistic conditions, including multiple graph topologies, limited adversarial visibility of the network, and clients with non-IID data. Finally, we show the limitations of existing defenses adapted from FL and design a new defense that successfully mitigates the backdoor attacks, without an impact on model accuracy. 

\end{abstract}


\section{Introduction}

Recently, machine learning (ML) has transformed a variety of real-life applications including self-driving cars
\iffullpaper
\else
\cite{car_2},
\fi
recommendation engines
\iffullpaper
\else 
\cite{recommendation_1},
\fi
 and personalized health 
\iffullpaper
\else 
\cite{health_2}.
\fi
A common thread to all these applications is data centralization, where a significant amount of data is collected for training ML models, a process that introduces risks to the privacy of users contributing their datasets. 
Several regulations such as the General Data Protection Regulation (GDPR) \cite{EUdataregulations2018} and the California Consumer Privacy Act (CCPA) \cite{CCPA} were specifically designed to protect data privacy.

To address the privacy concerns of centralized learning, McMahan et al.~\cite{mcmahan2017communication}, and Konecny et al.~\cite{Konecny} proposed Federated Learning (FL), a paradigm that trains ML models in a distributed fashion. In FL, personal data never leaves the device. Instead, devices individually update a  global model and share their model updates with a central server, which aggregates them and re-distributes the new global model to the clients.
Privacy  in FL can be enhanced with Multi-Party Computation (MPC)~\cite{bonawitz2017practical} and differential privacy\cite{brendan2018learning}, but most FL deployments do not utilize these technologies and are vulnerable to privacy attacks~\cite{NEURIPS2020_c4ede56b,DBLP:conf/icml/WenGFGG22}. 




In response to the privacy and reliability risks of FL with a single aggregation server, protocols for   Peer-to-Peer Federated Learning (P2PFL) have been proposed~\cite{NIPS2017_f7552665,hu2019decentralized,pmlr-v84-bellet18a, vanhaesebrouck2017decentralized, lalitha2019peer,fang2022bridge,yang2019byrdie,kuwaranancharoen2020byzantine,PENG2021108020}.
In P2PFL nodes communicate with their peers in the network and aggregate the model updates received from their peers. 
Multiple aspects of P2PFL have been studied, including adversarial settings (Byzantine~\cite{gupta2021byzantine} vs non-Byzantine~\cite{pmlr-d2}), the P2P network topology (complete graphs \cite{gupta2021byzantine} vs non-complete graphs \cite{lalitha2019peer}), and the output of the learning protocol (global  model learned by all peers\cite{gupta2021byzantine} vs. personalized models~\cite{pmlr-v84-bellet18a}). 


Real-life deployments of FL \cite{google_speech, apple_audio} raised concerns about adversarial attacks such as poisoning. While  such attacks have been extensively studied in centralized learning~\cite{biggio2012poisoning,mei2015using,xiao2015feature,koh2017understanding, chen2017targeted,suciu2018does, shafahi2018poison,jagielski2018manipulating,gu2019badnets}, 
they are more feasible in Federated Learning because adversaries can own or compromise mobile devices and participate in the FL training process. 
Attack vectors in FL include data poisoning~\cite{tolpegin2020data}, and model poisoning~\cite{BhagojiCMC19, bagdasaryan2020backdoor},  where the attacker aims to pursue availability, targeted, or backdoor attacks.
The objective of an availability attack is to decrease model accuracy and its utility~\cite{fang2020local, shejwalkar2021manipulating}, while targeted and backdoor attacks~\cite{sun2019can, wang2020attack} impact only a subpopulation of samples without dropping the overall model accuracy.


\revision{To the best of our knowledge, poisoning attacks in P2PFL have not been studied in the literature. As poisoning availability attacks result in degradation of model performance and can be detected through standard ML metrics, we focus on more sophisticated data poisoning attacks with stealthy objectives, in particular backdoor attacks in P2PFL.}
In addition, most P2PFL systems~\cite{gupta2021byzantine,  yang2019byrdie, fang2022bridge} either  consider that the P2P network topology is a complete graph or do not mention the  topology at all. Complete network topologies have inherent scalability issues and extremely high network bandwidth costs. In practical deployments of P2PFL, such connectivity assumptions are almost infeasible to satisfy, thus it is important to study attacks in P2PFL considering realistic network topologies. \revision{Compared to centralized FL, where a central aggregator receives all the malicious updates in each rounds and aggregates them into the global model, backdoor attacks are more challenging to mount in P2PFL as nodes only exchange updates with their neighbors and propagating malicious updates depends on network connectivity. }

In this paper, we propose and evaluate backdoor attacks in P2PFL where the adversary controlling a small set of nodes, has two goals: remaining stealthy, and achieving high attack success on samples with the backdoor pattern. These goals are conflicting and difficult to achieve simultaneously: as the attack becomes more successful it likely causes degradation in model accuracy, which can be detected by defenders.  We consider realistic graph topologies~\cite{
martins2010hybrid,jiang2019new,dong2015experimental,liu2012distributed} 
in which attackers can leverage information about the graph structure to increase the effectiveness of their  attack.  Specifically, our contributions are:

$\bullet$ We present a modular architecture for P2PFL that supports diverse network topologies and separates the learning graph from the communication graph to study poisoning attacks in realistic settings. 
Our current implementation uses GossipSub as the communication layer and P2P gradient averaging as the learning protocol. All code is publicly released\footnote{https://github.com/gokberkyar/BP2PFL}.
 
$\bullet$ We propose backdoor attacks in P2PFL and introduce new attack placement strategies based on graph centrality metrics such as node degree, PageRank, and clustering coefficient. We show that  a small number of attackers  (5\%) placed in the graph strategically, is sufficient to perform a backdoor attack with high attack success without decreasing the model accuracy for multiple graph topologies. 
We show that backdoor attacks can further be amplified by the attacker causing network failures that result in missed peer updates. We also demonstrate that an attacker with partial visibility into the network (e.g., 20\% of the nodes) can still successfully introduce a backdoor in the model. 
 
$\bullet$ Our paper is the  first extensive study on the impact of P2PFL backdoor attacks on various learning network topologies. 
Our study shows that the Barabasi-Albert scale-free network is the most vulnerable due to the presence of ``hubs'' (highly connected nodes). In a strategically placed attack, compromising the hub nodes will result in a particularly strong backdoor attack. 



$\bullet$  We introduce a new P2P defense based on weighting a node's contribution higher than the contributions of its peers when training each model. 
We propose a defense that uses two different clipping norms, one for peer updates, and one for local models, and thus bounds the contribution of a node's neighbors to  effectively prevent backdoor attacks.

$\bullet$ We analyze the impact of label imbalance in non-IID settings, where peers have different data distributions. We show that the non-IID model converges slower than IID to correctly classify clean data due to heterogeneity in local updates, however, the attack is still able to induce a high level of misclassification. 


\iffullpaper
\else 
\section{Threat Model}
\fi

\myparagraph{Adversarial goal}
An attacker may perform poisoning attacks with different goals, such as availability, targeted, or backdoor attacks. 
In this paper, we focus on backdoor attacks due to their particularly insidious nature and the feasibility of having malicious participants in P2P protocols. 
In backdoor attacks, the adversary has two goals: 1) Remaining stealthy, and 2) Achieving high attack success on backdoored samples. These objectives are conflicting and difficult to achieve simultaneously: as the attack becomes more successful at inducing misclassifications on backdoored samples, it likely causes degradation in model accuracy, which can be detected by defenders.  


\vspace{2pt}
\myparagraph{Adversary participation in the P2PFL protocol}
As P2PFL is an open system, we will assume the attacker either controls or compromises a number of $k$ peers, where $k<N$, with $N$ the total number of peers in the system. Compromising a peer in this context might be easier than compromising a well-protected aggregation server in FL.

\vspace{2pt}
\myparagraph{Attacker capabilities}
We assume that the attacker has full control over compromised peers. More precisely, the attacker can add, modify, or delete training data samples, modify and deviate from the machine learning algorithm, as long the final  model update has the same vector dimension as the model updates sent by  benign users. For example, the attacker can increase the number of local epochs used in training, change the learning rate of the model, or even  apply a new learning objective function. The attacker can only observe model updates received from the peers of compromised nodes, but no other nodes in the system.

\revision{
This threat model is motivated by common  design and assumptions in P2P architectures, in which it is  relatively easy for a party to become a participant in the P2P system. Thus, it is sufficient for the attacker to create several identities, i.e., mount a Sybil attack~\cite{10.1007/3-540-45748-8_24}), and add these multiple peers to the P2P system. These attacks have been shown to be feasible for many P2P systems such as streaming~\cite{multicast_ton_2008,pollution_streaming_2007},
DHTs \cite{secure_overlay_2003}, overlay communication \cite{pitn_icdcs_2016}, publish-subscribe \cite{gossipsub_sp_2024}, file systems such as IPFS \cite{ipfs_usenix_2022}, or general reputation systems \cite{hoffman2007survey}. Unfortunately, Sybil attacks are very difficult to defend against, but even when such defenses exist, peers can still be compromised and our attacks require a very small percentage of compromised nodes (5\%) to be successful.  
Our threat model is also consistent with the assumptions made in federated learning architectures that consider poisoning attacks~\cite{bagdasaryan2020backdoor, DBLP:conf/icml/BhagojiCMC19,sun2019can,wang2020attack}. In these settings, the attacker has also full control over the compromised clients, their local datasets, and can modify the model updates sent to the aggregation server by performing model boosting to increase the impact of the attack. }

In addition to controlling malicious clients, the attacker has network-level capabilities. In our strongest attack model, the attacker has full visibility of the peer connections and may use this knowledge to select and compromise the most critical nodes in the communication graph. 
Such powerful attackers are  relevant in  evaluating and comparing  mitigation strategies. We also consider a more relaxed adversarial model in which the adversary has partial visibility over the network, by observing a small fraction (e.g., 20\%) of the nodes in the communication graph.




\vspace{2pt}
\myparagraph{Attacker strategy}
Backdoor attacks have three components in P2PFL: pre-training phase, training time phase, and inference phase. In the pre-training phase, the adversary injects the desired backdoor (computed via standard methods~\cite{gu2017badnets}) into a subset of the training data at compromised peers. Specifically, if the attacker compromises peer $i$, has full access to $D_i$ and picks a Poisoning Data Ratio (PDR), then injects $\mbox{PDR} \cdot |D_i|$  backdoored samples into the training dataset. In the training phase, the adversary modifies the hyper-parameters of the local training and modifies model weights to conduct a model poisoning attack, e.g.,~\cite{bagdasaryan2020backdoor}. Finally, at inference time, if the adversary wishes to change the prediction of a certain sample, it injects the desired backdoor pattern.


\section{P2P Federated Architecture}
\label{sec:architecture}

In this section, we introduce the P2PFL architecture, the design decisions, and the Personalized Peer-to-Peer Averaging Algorithm.

\subsection{P2PFL Architecture Design}
\label{subsec:design}

\begin{figure}[!t]
    \centering
       \includegraphics[width=0.8\columnwidth]{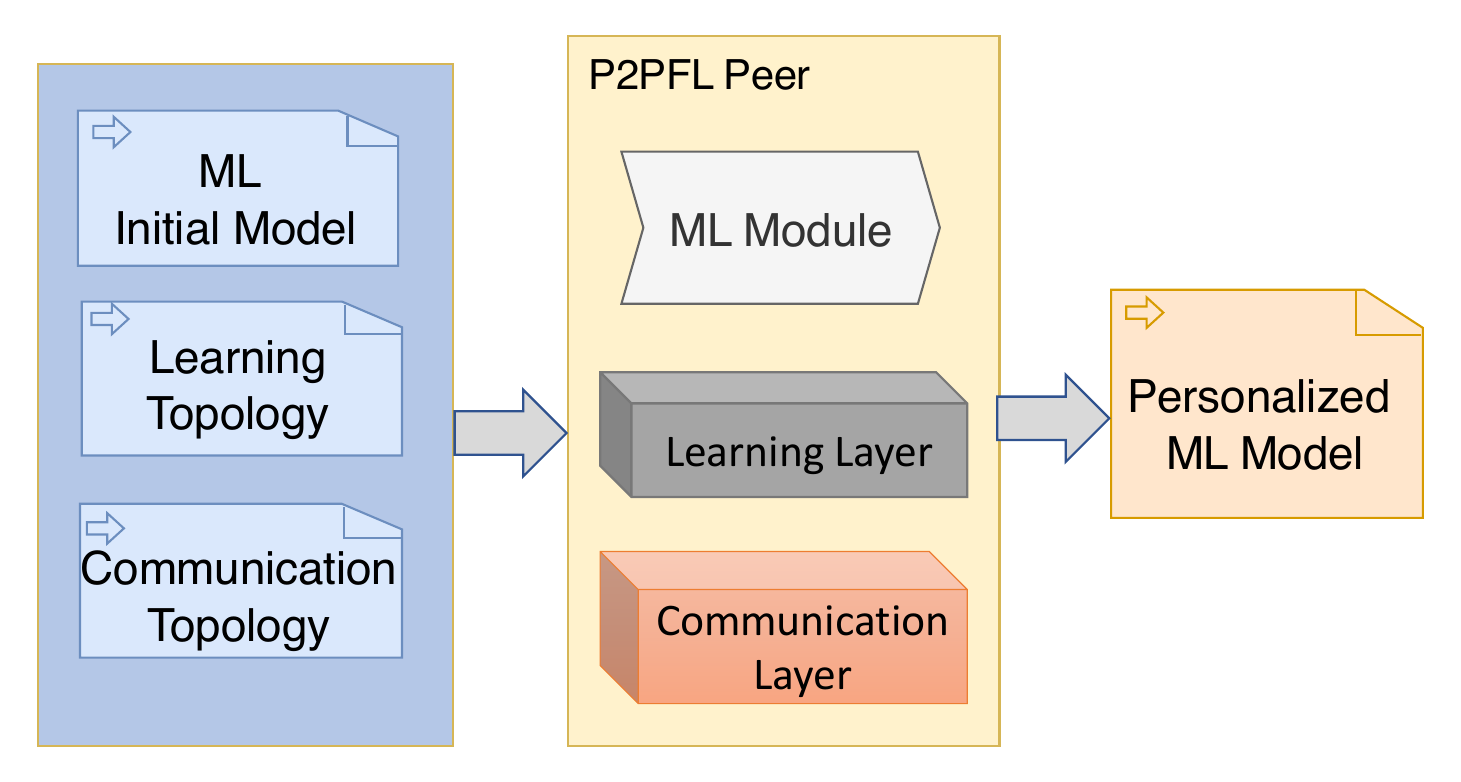}
           \caption{P2PFL Architecture Overview: 
            A peer has 3 roles: 1) Forwarding network packages (Communication Layer); 2) Sending and receiving ML updates to data peers (Learning  Layer); 3) Running ML training on  local dataset, aggregating updates received by the Learning Layer, and sharing back with the Learning Layer for the next round (ML Module).}
            \vspace{-10pt}
    \label{fig:Architecture}
\end{figure}

As previous  P2P learning systems assumed a connected graph for communication~\cite{gupta2021byzantine}, we consider realistic network topologies and create a modular architecture shown in Figure~\ref{fig:Architecture} that separates the learning and communication graphs. 

\begin{figure}[t]
    \centering
    \includegraphics[width=0.6\columnwidth]{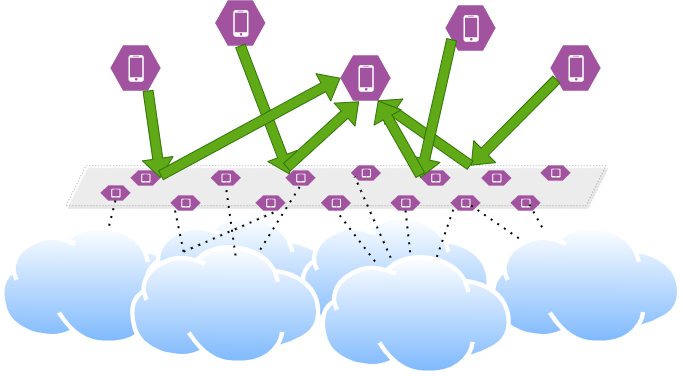}
        \caption{Communication topology is shown by the gray plane, learning topology uses the underlying communication topology to exchange updates. }
        \label{fig:platform}
        \vspace{-10pt}
\end{figure}

We define the \textit{learning topology} as a graph in which peers are connected if they exchange model updates during training. We call these peers \textbf{learning peers}.
We define the \textit{communication topology} as the graph in which peers  are connected if they can route network packets between them. The learning and communication topologies are shown in Figure~\ref{fig:platform}.
To distinguish the communication and learning topologies, we design P2PFL as an application that runs over the GossipSub protocol~\cite{gossipsub}. GossipSub constructs a mesh where nodes can publish their updates or subscribe to other nodes' updates, and has been shown to successfully balance excessive bandwidth consumption and fast message propagation~\cite{gossipsub}. When node A wants to communicate to node B, messages cannot be directly passed from A to B with a TCP connection, but rather A publishes its update into a channel that B subscribed into. We call these peers \textbf{communication peers}. This design choice provides better network bandwidth utilization and robustness to network failures. 
 
\subsection{P2P Gradient Averaging Algorithm}
\label{subsec:P2PFL_algorithm}

While our architecture  supports different ML algorithms, we 
use P2P Gradient Averaging (Algorithm~\ref{alg:personalized_p2p_averaging}), based on the standard Federated Averaging algorithm \cite{mcmahan2017communication,kairouz2021advances}. \revision{Variants of this training method have been proposed in the literature under the name of ``decentralized learning''~\cite{NIPS2017_f7552665,hu2019decentralized,pmlr-v84-bellet18a, vanhaesebrouck2017decentralized, lalitha2019peer,fang2022bridge,yang2019byrdie,kuwaranancharoen2020byzantine,PENG2021108020}. As there are many published variants of decentralized learning, we select Algorithm~\ref{alg:personalized_p2p_averaging} as a representative P2PFL method.}
In this method, each node trains a personalized model using the  updates received from its neighbor set in the learning graph.

\begin{algorithm}
\small

\KwData{Local Dataset $D$, rounds $T_N$, peer set $S$}

\SetKwFunction{Diff}{P2PAverage}
\SetKwProg{Fn}{Function}{:}{}


\Fn{\Diff{}}{
    $f_0 = \textsc{GetInitialModel}(0)$ 
    
    
    \For{$t \in [1, T_N]$}{
        \textcolor{gray}{// Compute local update using SGD}
        
        $A^t=\textsc{ComputeLocalUpdate}(f_{t-1}, D)$

        \textcolor{gray}{// In parallel send and receive updates}
        
        $U^t=\textsc{GetUpdates}(S)$

        $\textsc{SendUpdate}(A^t, S)$

        $f_t = \textsc{Aggregate}(A^t, U^t)$ 
      
    }
    
    \Return $f_n$
}
\caption{Personalized Peer-to-Peer Gradient Averaging}
\label{alg:personalized_p2p_averaging}
\end{algorithm}

Specifically, each node participating in the protocol owns its private training dataset $D$, a set of peer nodes $S$, to which the node can exchange model updates and its personalized model $f_t$ at round t. Each node synchronously trains its personalized model $f_t$ at round $t$ by computing its local update $A^t$ using its private dataset $D$, and then aggregating with the previous round's personalized model $f_{t-1}$, and model updates received from its peers $U^t$. For the aggregation step, we average all the model updates received from the neighboring peers. The framework supports the use of more advanced Byzantine robust aggregation functions such as KRUM~\cite{KRUM} and Bulyan~\cite{Bulyan} that have been proposed for poisoning defense in FL. We adapt the Trimmed  Mean~\cite{xie2018phocas} and gradient clipping~\cite{bagdasaryan2020backdoor, sun2019can} FL defenses to the P2PFL setting. 



\section{Backdoor Attacks on P2P Federated Learning}
\label{sec:attacks}
In this section we describe the backdoor attacks we propose for P2PFL and discuss different attack strategies that select relevant peers based on graph centrality metrics.


\subsection{Backdoor Attack on P2PFL}


We assume the adversary has full access to the peer’s private training data, model weights, and other training parameters in each compromised peer and  runs a backdoor attack on each compromised peer using  
\iffullpaper
\else
Algorithm~\ref{alg:backdoor_node}. 
\fi
The attacker  uses the BadNets attack by Gu et al.\cite{gu2017badnets}, amplified by a model poisoning attack \cite{bagdasaryan2020backdoor,sun2019can}. The attacker chooses a Poisoning Data Ratio and the Boosting Factor to amplify the contribution of the local model. Increasing the Poisoning Data Ratio and Boosting Factor increases the attack success, but also causes a drop in test accuracy  as a side effect. 
\begin{algorithm}
\small

            

\KwData{Target Node $target$, Local Dataset $D$, rounds $T_N$, peer set $S$, poison data rate $PDR$, boosting factor $B$, target class $C_t$}

\SetKwFunction{Diff}{AttackSingle}
\SetKwProg{Fn}{Function}{:}{}


\Fn{\Diff{$target, PDR, B, C_t$}}{
    $f_0 = \textsc{GetInitialModel}(0)$ 

    $D^* = \textsc{BackdoorDataset}(D, PDR, C_t)$

    
    \For{$t \in [1, T_N]$}{
        \textcolor{gray}{// Compute local update on Backdoored Dataset}
        
        $A^t=\textsc{ComputeLocalUpdate}(f_{t-1}, D^*)$
        
        \textcolor{gray}{// Do model poisoning attack by boosting}
        
        $A^{t^*}=\textsc{ModelPoisonUpdate}(A^t, B)$

        \textcolor{gray}{// In parallel send and receive updates}
        
        $U^t=\textsc{GetUpdates}(S)$
        
        \textcolor{gray}{// Send malicious  updates}
        
        $\textsc{SendUpdate}(A^{t^*}, S)$

        $f_t = \textsc{Aggregate}(A^{t^*}, U^t)$ 
      
    }
    
}

\caption{P2P Backdoor Attack Single Node}
\label{alg:backdoor_node}
\end{algorithm} 

\subsection{Stronger Structural Graph Attacks}
\label{sec:identification_critical}

The P2PFL protocol runs on a non-complete graph topology, thus not all peers have an equal impact  when used by the attacker. One natural question that the adversary must answer is how to choose the $k$ peers for attack, among all $N$ peers available in the system. For example, if the attacker has  budget for compromising only 5\% of all nodes, how can the adversary  maximize its adversarial goal by carefully selecting those attacker nodes?

One simple strategy for the attacker is to select the adversarial nodes randomly. However, considering the
P2P nature of the system, a natural approach is to select nodes that are well connected in the graph and have high centrality measures.
We introduce several attack strategies based on 
four  well-known graph centrality  metrics: maximum degree, Effective Network Size \cite{borgatti1997structural}, PageRank scores~\cite{page1999pagerank}, and maximum clustering coefficient~\cite{saramaki2007generalizations}. 

\vspace{2pt}
\myparagraph{Maximum degree} Nodes with the highest degree in the graph allow  the attacker to propagate malicious updates to a large number of neighbors.

\vspace{2pt}
\myparagraph{ENS score} 
\iffullpaper
\else
The  effective network size (ENS) of a node’s ego network can be computed as:
\fi
\begin{equation}
e(u) = n - 2t/n
\end{equation}
where $t$ is the number of ties in the ego network (not including ties to the node itself) and $n$ is the number of nodes in the ego network.
A recent work on cyber network resilience against self-propagating malware~\cite{chernikova2022cyber} observed that nodes with the highest ENS tend to act as bridges between two dense clusters and monitoring them prevents attack spreading. Our insight is that instead of using ENS for robustness, we select nodes with largest ENS scores to enable the attacker to traverse bridges in the network and compromise different clusters. 

\vspace{2pt}
\myparagraph{PageRank score} The PageRank score computes a ranking of the nodes in graph $G$ based on the structure of the incoming links~\cite{page1999pagerank} and 
provides a metric of centrality and node importance  in the graph. 
As PageRank  showed empirical success on sparse graphs and the attacker's goal is to identify critical nodes, we leverage nodes with highest PageRank score as a viable attack selection strategy.

\vspace{2pt}
\myparagraph{Maximum clustering coefficient}
For unweighted graphs, the clustering coefficient of a node  is the ratio of possible triangles through that node:
\begin{equation}
c_u = \frac{2 T(u)}{\mathsf{deg}(u)(\mathsf{deg}(u)-1)}
\end{equation}
where $T(u)$ is the number of triangles through node $u$ 
and $\mathsf{deg}(u)$ is the degree of node $u$~\cite{saramaki2007generalizations}. 
We select this metric as nodes with highest clustering coefficient tend to have a more connected local neighborhood where the malicious updates can propagate. 




\subsection{Quantifying Attack Success}
\label{sec:quantify_attack}
The attacker has two objectives: 1) Performing a successful backdoor attack, and 2) Remaining stealthy so that the attack is not detected by monitoring the model's accuracy. We quantify these goals with two metrics used in the poisoning literature: (1) {\bf Attack Success}, denoting the fraction of poisoned samples that were incorrectly classified as belonging to attacker's target class, and (2) {\bf Test Accuracy}, representing the fraction of clean samples that were correctly classified. These  two metrics are averaged across all participants. 
To evaluate the attack we use an auxiliary test set partitioned into two non-overlapping subsets: a clean dataset, and a backdoored dataset, on which we compute the test accuracy and attack success, respectively.



\section{Backdoor Attack Evaluation}
\label{sec:evalution}

In this section we evaluate the effectiveness of the attacks presented in Section~\ref{sec:attacks}. We first introduce our experimental setup,
then evaluate the effectiveness of attacks under several settings, by seeking to answer the following questions:

\begin{itemize}
\item What node selection strategy provides most benefit to the adversary?
\item What graph topology is more impacted by attacks?
\item How does compromising more peers affect the attack?
\item What is the impact of link failures on backdoor attacks?
\item How does the data distribution impact the attack?
\item What is the effect of constraining the adversary to a limited view of the network?

\end{itemize}

\subsection{Experimental Setup}
\label{sec:Experiment_Setup}

\myparagraph{Network topology} 
We study three representative network topologies, which have been widely used to model complex networks: (i)  Random graphs (Erdos-Renyi)~\cite{erdos1960evolution}; (ii) Small-world graphs (Watts-Strogatz)~\cite{watts1998collective}; and (iii) Scale-free graphs (Barabasi-Albert)~\cite{albert2002statistical}.

Small-world graphs are characterized by a small average path length and high clustering coefficient, properties that have been observed in real-world networks~\cite{newman2003structure}. In both random and small-world networks, nodes have comparable degrees, and, thus, the average degree can be viewed as the ``scale'' of the network. In contrast, in scale-free networks, the fluctuations from the average degree are large, with a few highly connected nodes serving as ``hubs'', while the vast majority have low degrees. The Internet is an example of a scale-free network, where the degree distribution is shaped by the ``preferential attachment'' to a small number of popular hubs~\cite{albert2002statistical}.

Table~\ref{table:topology} summarizes the parameters of the three topologies used in this study for a 60-node size network. We also experimented with smaller and larger networks from 30 to 100 nodes.

\begin{table}[!htbp]
    \centering
\begin{tabular}{|l|r|r|r|r|}
\hline
{\bf Parameter} &  {\bf Erdos R.} &  {\bf Watts S.} &  {\bf Barabasi A.} &  {\bf Complete} \\
\hline
\# Nodes         &        60.00 &                                 60.00 &            60.00 &     60.00 \\
\# Edges         &       166.00 &                                360.00 &           576.00 &   1770.00 \\
Mean Degree      &         5.53 &                                 12.00 &            19.20 &     59.00 \\
Density          &         0.09 &                                  0.20 &             0.33 &      1.00 \\
Diameter         &         5.00 &                                  3.00 &             3.00 &      1.00 \\
Radius           &         3.00 &                                  2.00 &             2.00 &      1.00 \\
Mean Distance    &         2.56 &                                  1.88 &             1.68 &      1.00 \\
Transivity       &         0.10 &                                  0.24 &             0.39 &      1.00 \\
Clustering coefficient &         0.09 &                                  0.24 &             0.41 &      1.00 \\
\hline
\end{tabular}
\caption{Characteristics of network topologies used in this study.}  
\label{table:topology}
\vspace{-10pt}
\end{table}

\vspace{3pt}
\myparagraph{Datasets} We used the MNIST~\cite{deng2012mnist} (60,000 samples), FashionMNIST~\cite{xiao2017fashionmnist} (70,000 samples), and  EMNIST Digits~\cite{cohen2017emnist} (270,000 samples) datasets, all featuring $28 \times 28$ pixel images labeled to one of 10 classes.  We also used a more complex image classification dataset, CIFAR-10, with $32 \times 32$ pixel images, 60,000 samples, and 10 classes. The partitioning method among peers has a large impact on the ML model, generating peer datasets that fall into two broad categories: independently and identically distributed (IID), and non-independent and identically distributed (non-IID). Non-IID data distribution is a common challenge in FL~\cite{li2022federated}.
In our paper, we analyze the attack performance in both IID and non-IID settings. In the IID setting, each client receives an equal number of samples of each label, while in the non-IID setting, each client is allocated a proportion of the samples of each label according to the Dirichlet distribution~\cite{pmlr-v97-yurochkin19a}. 

\vspace{3pt}
\myparagraph{Parameters} 
The attack is characterized by the following parameters: $k$, the number of malicious peers; PDR, the ratio of poisoned data over total samples in a malicious peer; boosting factor used in the model poisoning attack; adversarial training epoch count; and target class. 
We experimented with multiple values of these parameters. Here, we discuss the attack impact for various numbers of adversaries ($k$), while fixing the PDR to 0.5, the boosting factor to 10,  adversarial training epoch count to 5, and the target class to 2.
Results shown for various strategies and parameters are averaged over three different runs.

\vspace{3pt}
\myparagraph{Default configuration} Unless otherwise specified, the experiments use the PageRank selection strategy on Watts Strogatz 60-node topology with 5\% adversarial nodes. Furthermore, our default configuration uses the EMNIST dataset. Each peer receives a total of 5200 samples during training, with an equal number of samples per class (IID distribution).

\vspace{3pt}

\revision{
\noindent \myparagraph{Computing platform}
The experimental setup for simulating the P2P environment utilizes a virtual network of interconnected Docker containers, all running concurrently on a single machine. Each container is configured to act as an independent peer within the P2P network, thereby enabling a realistic simulation of interconnected devices. The communication between peers is achieved with GossipSub, a publish-subscribe system used by many P2P applications including Ethereum.
Our hardware is equipped with an Intel Xeon
Silver 4114 CPU with 32 cores and 192GB of RAM. In the default configuration of the experiments, we run a network of 60 interconnected Docker containers, and we change the number of nodes to 30 in some of the experiments.
}

\subsection{Adversary's Node Selection Strategy}
\label{sec:selection} 

\begin{figure}[!ht]
    \centering   
    \vspace{-10pt}
       \subfloat[][Test Accuracy]{%
       \includegraphics[width=0.48\linewidth]{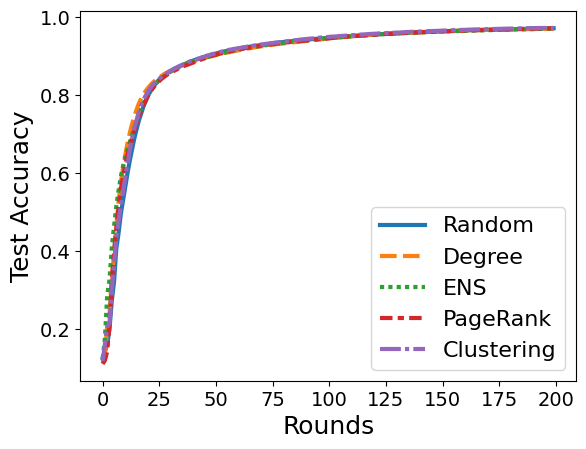}
       \label{fig:selection_acc}
       }
       \subfloat[][Attack Success]{%
    \includegraphics[width=0.48\linewidth]{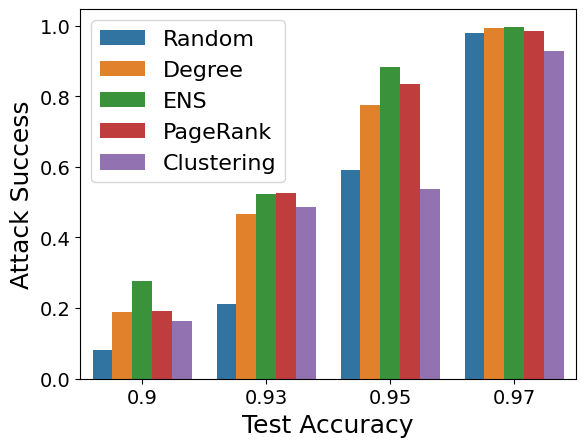}
       \label{fig:selection_attack}
       }
       \caption{Adversary's node selection strategy: (a) accuracy of the backdoored model on clean test data. (b) attack success (y-axis) of various strategies after the backdoored model has reached high accuracy (x-axis).} 
    \label{fig:selection}
\end{figure}

We evaluate the success of the attack for the node selection strategies introduced in Section~\ref{sec:identification_critical}: Random, Degree, ENS, PageRank, and Clustering.
Figure~\ref{fig:selection} shows the evolution of P2PFL model's performance across training rounds. 
The test accuracy of the backdoored model on clean data converges similarly for all five attack strategies, and eventually reaches 0.97. In practice, the training may stop after a certain accuracy has been reached. Therefore, in Figure~\ref{fig:selection_attack} we evaluate the attack success at fixed test accuracies. We notice a significant difference between the four attack strategies for accuracy in the low 90s (i.e., 0.9, 0.93), with the centrality-based methods Degree, ENS and PageRank being consistently on top, generally twice more successful than Random at misclassifiying the adversarial test samples.
Once the model converges to 0.97 accuracy, the attack is highly successful regardless of the node selection method. We observe tough that as the training runs for longer to convergence and the test accuracy is higher, the backdoor is also learned better by the model and the attack becomes more successful for all selection strategies. This phenomena demonstrates the well-known tradeoff between model performance (measured via accuracy) and attack success.

\subsection{Network Topology}
\label{sec:topo}

\begin{figure}[!t]
    \centering   
    \subfloat[][Test Accuracy]{%
       \includegraphics[width=0.48\linewidth]{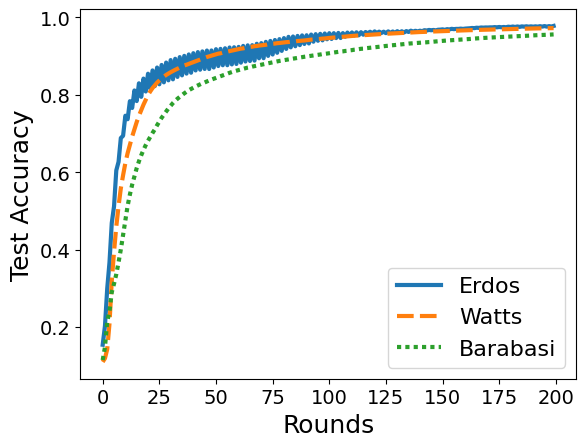}
       \label{fig:topology_acc}
       }
       \subfloat[][Attack Success]{%
    \includegraphics[width=0.48\linewidth]{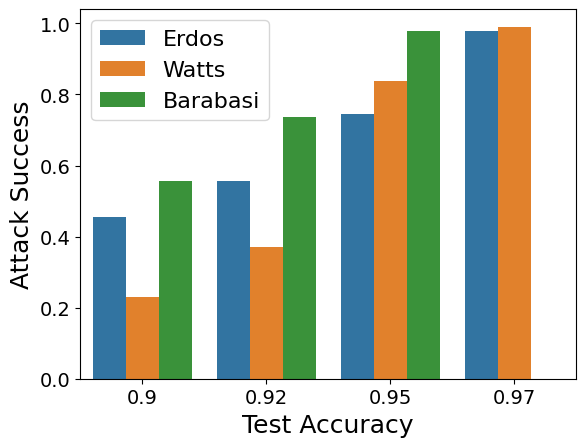}
       \label{fig:topology_attack}
       }
       \caption{Impact of network topology on attack performance: (a) accuracy evolution; (b) attack success (y-axis) on various topologies after the backdoored model has reached high accuracy (x-axis). Within 200 rounds of training, the Barabasi topology has not exceeded 0.95 test accuracy (hence, the figure omits Barabasi at 0.97 accuracy).} 
    \label{fig:topology}
    \vspace{-10pt}
\end{figure}

We study the impact of the network topology on the attack. We compare the three types of graphs described in Table~\ref{table:topology}:  Erdos Renyi, Watts Strogatz, and Barabasi Albert, under a PageRank-based attack strategy. These graph models are typically used to analyze the behavior of social media and cellular network graphs, with previous work suggesting that P2PFL networks will most likely be small-world networks~\cite{
martins2010hybrid,jiang2019new,dong2015experimental,liu2012distributed}.
Figure~\ref{fig:topology_acc} presents the accuracy of the backdoored P2PFL model across multiple rounds of training, and
Figure~\ref{fig:topology_attack} illustrates the attack success after the model has reached high accuracy ($\geq 0.9$) on clean data. 
These results point out a major insight: Barabasi scale-free network is the most vulnerable, due to the presence of highly connected nodes (hubs) that are selected with PageRank as the target of attack. The backdoored model becomes highly successful within 200 rounds of training on all topologies, learning to classify clean samples correctly (accuracy $\geq$ 0.95), but also to misclassify poisoned samples (0.99 attack success).

\subsection{Scaling Up the Attack}
\label{sec:scale}  

\begin{figure}[!h]
    \centering   
    \vspace{-10pt}
    \subfloat[][Adversarial Nodes]{%
       \includegraphics[width=0.46\linewidth]{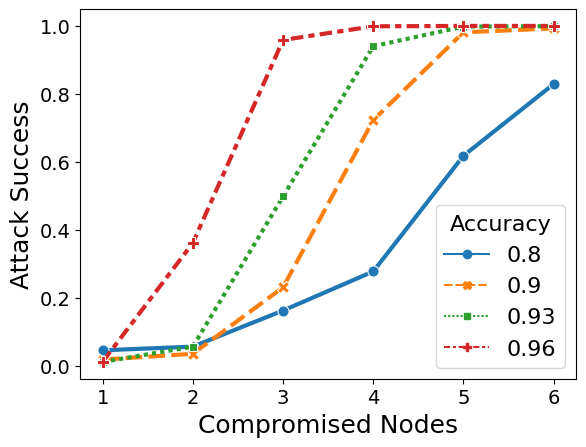}
       \label{fig:scale_adversaries}
       }
       \subfloat[][Fault Tolerance]{%
    \includegraphics[width=0.46\linewidth]{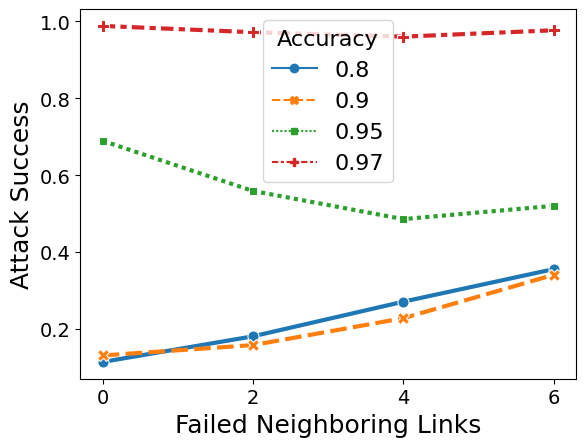}
       \label{fig:scale_fault}
       }
       \caption{(a) Increasing the number of compromised nodes within a 60-node network. (b) Increasing the number of failed connections to neighbors (i.e., missed updates), with 3 compromised nodes.} 
    \label{fig:scale}
    \vspace{-10pt}
\end{figure}

Figure~\ref{fig:scale_adversaries} shows that the attack scales well with the number of adversarial nodes.
Given a desired test accuracy, (i.e., 0.8, 0.9, 0.93, 0.96 in the figure), we evaluate the attack success for 1 to 6 compromised nodes selected with the PageRank strategy, on the 60-node Watts Strogatz topology. Note that our previous experiments have used only 3 adversarial nodes (i.e., 5\%) while still delivering high performance. 
We also analyzed scaling to larger networks of 80 and 100 nodes, with expected results (not shown in figure): the convergence speed slightly decreases as the system scales up, however the attack is still highly successful.

\subsection{Fault Tolerance}
\label{sec:fault_tolerance} 

As P2PFL is a collaborative distributed system, node failures are inevitable, and any P2PFL system should be tolerant to failures. 
In our study, we assume failures affect random nodes, and result in missed peer updates that do not contribute to the learned model.
Figure~\ref{fig:scale_fault} measures the impact of 0, 2, 4, and 6 failed neighboring links per peer. Missing updates is generally in attacker's favor at lower accuracy of 0.8 and 0.9. As the model learns to classify clean data better (test accuracy of 0.95 and 0.97) and the attack becomes more successful, missed updates do not improve the attack any longer. 

\subsection{Impact of Dataset}
\label{section:Dataset}

In the previous experiments, we have shown that the  P2P backdoor attack is highly successful on the EMNIST dataset. In this section, we explore its transferability to other datasets: MNIST, Fashion MNIST \revision{and CIFAR-10}. These latter datasets are about $5\times$ smaller than EMNIST, therefore we reduce the number of peers to 30. We ensure that training is carried out on the same number of samples (based on the size of the smallest training dataset, MNIST), i.e., 1800 samples per client.

\begin{figure}[ht]
    \centering   
    \vspace{-10pt}
    \subfloat[][Test Accuracy]{%
       \includegraphics[width=0.48\linewidth]{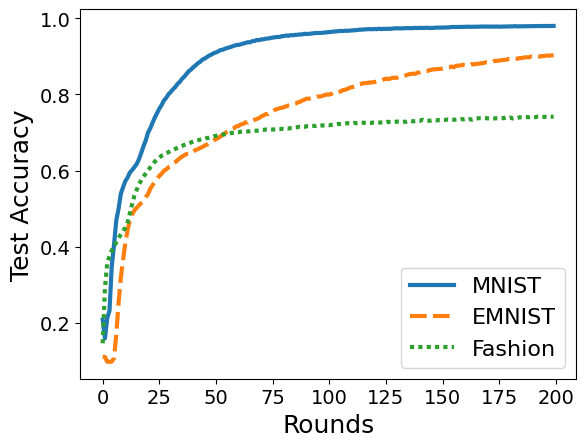}
       \label{fig:dataset_acc}
       }
       \subfloat[][Attack Success]{%
    \includegraphics[width=0.48\linewidth]{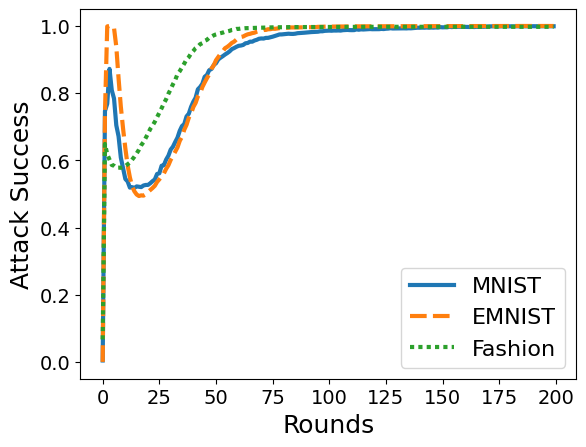}
       \label{fig:dataset_attack}
       }
       \caption{Datasets comparison: (a) accuracy and (b) attack success of the backdoored P2PFL model. Configuration: 30 peers, 3 malicious nodes, Watts-Strogatz topology with PageRank attack.} 
    \label{fig:dataset}
\end{figure}

Figure~\ref{fig:dataset_acc} presents the evolution of accuracy while training the P2PFL backdoored model. We observe that the model converges at different speeds for the various datasets. After 200 rounds of training, Fashion MNIST reaches 0.76 accuracy, EMNIST 0.91 and MNIST 0.98. Fashion MNIST is a more complex learning task than MNIST and EMNIST, and thus the accuracy is lower. 
We further note that the PageRank-based attack is highly successful at misclassifying poisoned samples on all datasets (Figure~\ref{fig:dataset_attack}), demonstrating that backdoor attacks are a valid threat for multiple applications. 
We notice an early peak in attack success when accuracy is still very low, indicating that malicious nodes (whose impact is amplified by gradient boosting) are more effective in the first rounds, before their contribution is offset by the honest majority.
Reducing the boosting factor (we experimented with values of 10, 5, and 1) or the number of malicious peers has the effect of decreasing this early peek. As in the previous experiments on EMNIST, the attack success increases as more rounds of training are performed and the model converges.

\revision{\myparagraph{Complex Datasets} To demonstrate the generality of the attack and its applicability to more complex tasks we evaluate our method over the CIFAR-10 dataset. The CIFAR-10 dataset consists of 60,000 32x32 samples representing colored images. Each sample belongs to one of 10 classes, with 6000 images per class. The samples are split between 50,000 training images and 10,000 test images. The model used for the experiments is a CNN with 551,466 parameters and six convolutional layers. We split the data uniformly at random to 30 clients, and consider 3 malicious clients and a data poisoning rate of $0.5\%$.  Due to the increased difficulty of the learning task each experiment took approximately 19 hours to finish training for 200 rounds.} 

\revision{
To accommodate the requirements of this more complex dataset we changed
our implementation to handle larger messages sent between the peer by chunking these messages. Additionally, due to the artifact that we ran all our experiments on one physical machine that is not equipped with GPUs and to maintain consistency with the other experiments we use
the aforementioned six-convolution architecture and we considered a deployment on 30 nodes. This is also consistent with other deployments of the CIFAR-10 model in a federated setting where a similar amount of nodes are used \cite{koloskova2020decentralized}, \cite{daily2018gossipgrad}, \cite{he2023byzantinerobust}.}

\begin{figure}[ht]
    \centering   
    \vspace{-10pt}
    \subfloat[][Test Accuracy]{%
       \includegraphics[width=0.48\linewidth]{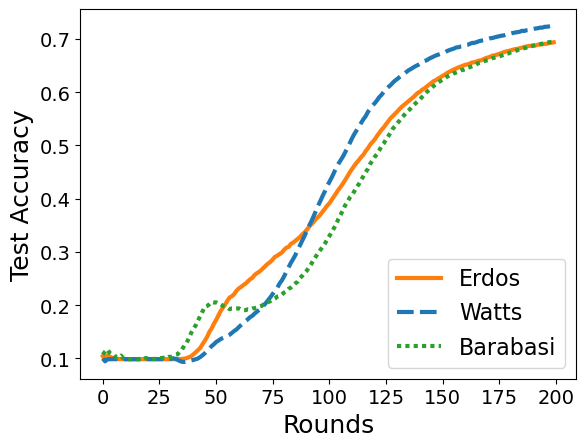}
       \label{fig:cifar10_acc}
       }
       \subfloat[][Attack Success]{%
    \includegraphics[width=0.48\linewidth]{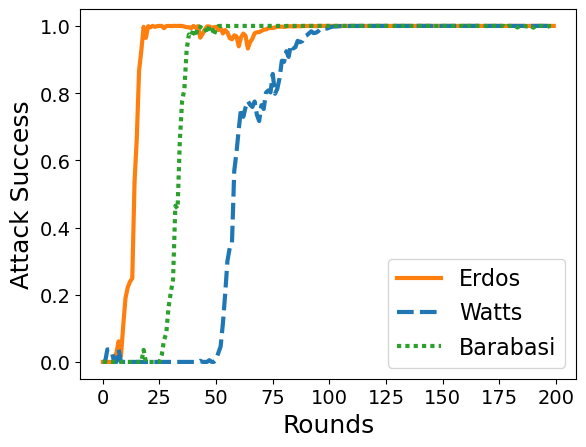}
       \label{fig:cifar10_attack}
       }
       \caption{\revision{Impact of network topology on attack performance on CIFAR-10: (a) accuracy evolution; (b) attack success over 200 rounds.}}
    \label{fig:cifar10_dataset}
\end{figure}

\revision{Figure~\ref{fig:cifar10_acc} illustrates the accuracy evolution during the training of the P2PFL backdoored models on CIFAR-10, using three graph topologies: Erdos, Watts Strogatz, and Barabasi. Given the increased complexity of CIFAR-10 compared to MNIST, EMNIST, and Fashion MNIST datasets, early convergence is challenging, requiring approximately 25 rounds before the exchanged gradient updates significantly impact the models' convergence. After 200 rounds, each model achieves around 70\% accuracy for all graph types. Additionally, our attack method effectively misclassifies poisoned samples regardless of the network graph topology with the attack reaching almost 100\% after 100 rounds (Figure~\ref{fig:cifar10_attack}).  However, the system demonstrates increased resilience with the Watts Strogatz graph topology, as the attack requires more epochs of training to succeed. Conversely, the attack is quicker to converge with the Erdos and Barabasi graph topologies, though this results in a reduction in test accuracy. }

\begin{figure}[ht]
    \centering   
    \vspace{-10pt}
    \subfloat[][Test Accuracy]{%
       \includegraphics[width=0.48\linewidth]{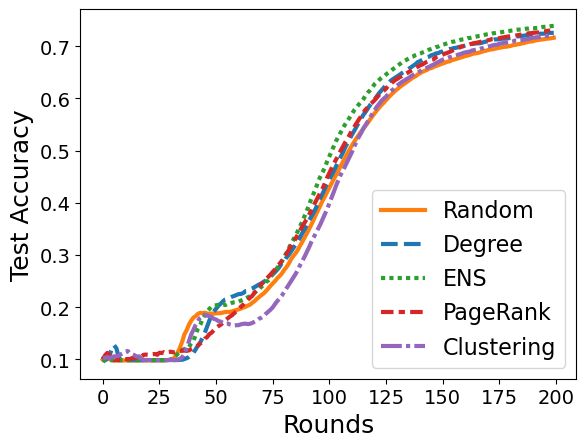}
       \label{fig:cifar10_selection_acc}
       }
       \subfloat[][Attack Success]{%
    \includegraphics[width=0.48\linewidth]{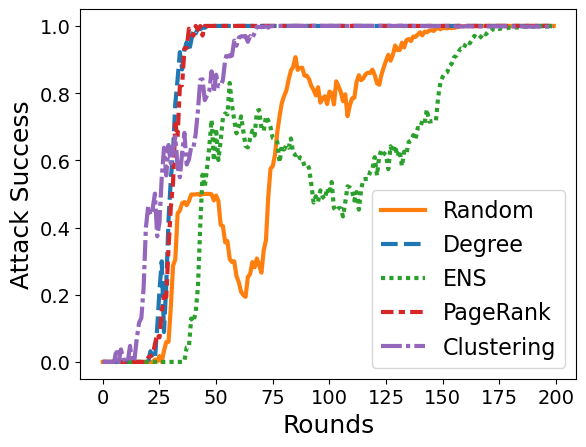}
       \label{fig:cifar10_selection_attack}
       }
       \caption{\revision{Adversary's node selection strategy with CIFAR-10: (a) average accuracy of the backdoored models on clean test data. (b) attack success (y-axis) of various node selection strategies after 200 rounds.}}
    \label{fig:cifar10_selection}
\end{figure}

\revision{In Figures \ref{fig:cifar10_selection_acc} and \ref{fig:cifar10_selection_attack}, we show the test accuracy and backdoor attack success for several adversarial node selection strategies: degree, ENS, PageRank, clustering coefficient, and a random baseline on the Watts Strogatz graph. All strategies impact minimally the models' test accuracy, as the attack is stealthy. We  note that different adversarial node selection strategies impact the attack's success rate, primarily influencing the convergence speed. For example, the random selection strategy is less efficient than selecting peers based on the highest degree. All in all, the PageRank selection strategy proves to be optimal, as it minimally impacts test accuracy and allows the attack to converge efficiently. Finally, it is important to note that the attack achieves a 100\% success rate differently across datasets. A combination of a low Data Poisoning Rate (0.5\% for CIFAR-10 and 50\% for MNIST-like datasets) and the difficulty of the learning task makes the attack on CIFAR-10 stealthier and more effective, as the benign gradients cannot overwrite the malicious gradients. Thus, no 'dip' is observed in the attack success curves, and a steady increase is maintained instead.}
 

\subsection{Impact of Data Distribution}
\label{section:iid_vs_noniid}

We next compare the attack's performance (PageRank strategy) in IID settings (where nodes have an equal number of samples of each label), against non-IID settings (where nodes are allocated different numbers of samples of each class). We model the label imbalance of non-IID using the Dirichlet distribution~\cite{pmlr-v97-yurochkin19a}, which is a common choice for simulating real-world data partitioning~\cite{Wang2020FedNova,li2022federated}.  The Dirichlet distribution is denoted $Dir(\alpha)$. The concentration parameter $\alpha$ controls the degree of similarity between peers. As $\alpha \rightarrow \infty$, peers' distributions become identical, whereas as $\alpha \rightarrow 0$, distributions become extremely imbalanced, with each class residing on separate peers. 

\begin{figure}[!th]
    \centering   
    \vspace{-15pt}
    \subfloat[][Test Accuracy]{%
       \includegraphics[width=0.46\linewidth]{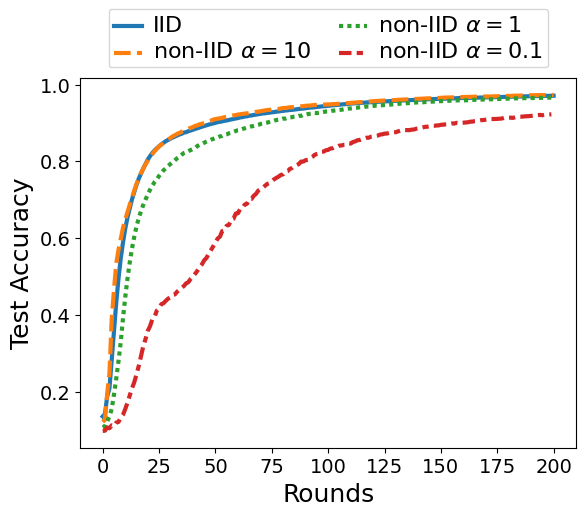}
       \label{fig:distrib_acc}
       }
    \subfloat[][Attack Success]{%
    \includegraphics[width=0.46\linewidth]{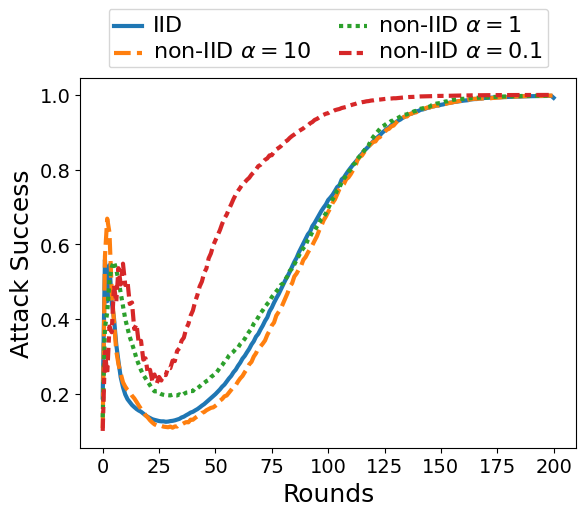}
       \label{fig:distrib_attack}
       }
       \caption{Impact of data distribution: IID and non-IID settings for three values of the concentration parameter $\alpha$: 10, 1, and 0.1. 
       } 
    \label{fig:distrib}
    \vspace{-5pt}
\end{figure}

Figure~\ref{fig:distrib} compares IID against non-IID for three values of $\alpha$: 10, 1, and 0.1. We observe that learning becomes more challenging in non-IID settings as the class imbalance increases (i.e., for smaller $\alpha$). Heterogeneity in the peers' local datasets leads to large variations in local updates performed by peers~\cite{li2022federated}. As a results, accuracy convergences slower (Figure~\ref{fig:distrib_acc})  and the attack is more successful in inducing misclassifications (Figure~\ref{fig:distrib_attack}). 
In non-IID settings, where the class distribution is more skewed towards a subset of peers, we end up with fewer honest peers holding enough correctly labeled samples of each class. Thus, the honest updates are overpowered by the boosted adversarial updates, and a correct prediction for samples belonging to the target class is more difficult to learn (see non-IID $\alpha=0.1$ from Figure~\ref{fig:distrib_attack}).


\subsection{Constrained Adversary with Partial View of the Graph}
\label{section:view}

In this section, we evaluate a constrained adversarial strategy, in which the attacker's view is restricted to a subset of the network. 
In our analysis, the observable subgraph represents 20\% of the nodes in the network. The initial visible node is randomly chosen, while other nodes are added to the observable subset based on an exponential decay formula, $p^d$, were $p$ is a probability parameter and $d$ is the depth (i.e., number of hops) from the initial node. In our experiments, nodes that are further than 3-hops away ($d > 3$) have a zero probability of joining the observable subset ($p = 0$), otherwise $p = 0.5$. The attacker is restricted to applying his node selection strategies to the observable subset.
\begin{figure}[!th]
    \centering 
    \subfloat[][Test Accuracy]{%
       \includegraphics[width=0.46\linewidth]{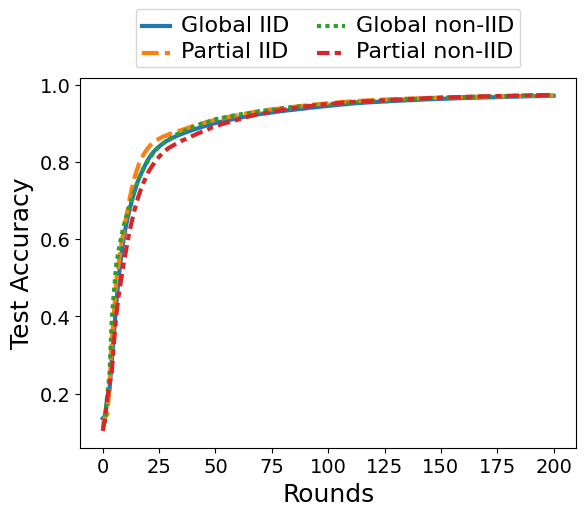}
       \label{fig:view_acc}
       }
     \subfloat[][Attack Success]{%
      \includegraphics[width=0.46\linewidth]{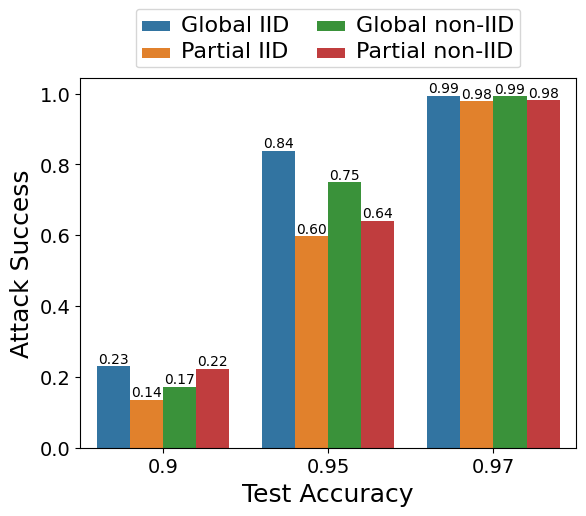}
       \label{fig:view_attack}
       }
       \caption{Constraining the adversary's view. We compare the Global with the Partial View in IID and non-IID ($\alpha=10$) settings. (a) test accuracy on clean data; (b) attack success after the backdoored model has reached high accuracy.} 
    \label{fig:view}
   \vspace{-10pt}
\end{figure}

Figure~\ref{fig:view} compares the constrained attacker (Partial view) against an attacker with full view of the network (Global view) under the PageRank-based attack. 
We observe that accuracy on clean test data converges similarly, regardless of attacker's view (Figure~\ref{fig:view_acc}), and reaches 0.97 in all cases.
Figure~\ref{fig:view_attack} analyzes the attack success of the backdoored P2PFL model at high accuracy levels ($\geq 0.9$). For IID, the Global view consistently achieves higher attack performance compared to the Partial view. However, non-IID settings (illustrated for concentration parameter $\alpha=10$) present a high variability, with the data distribution having a stronger impact on attacker's success than the observability restriction. Once the model has converged, non-IID and IID achieve the same level of attack success: 0.99 with Global view and 0.98 with Partial view.

\section{Backdoor Defense \label{sec:defense}}
We  first discuss standard defenses against  poisoning attacks in FL and show empirical evidence that these defenses are ineffective, achieve slower convergence, and reduce the model accuracy. We then propose  a new defense against poisoning in  P2PFL and show experimentally that our defense  counteracts backdoor attacks without a significant drop in model accuracy.

\iffullpaper
\else 
\subsection{Existing Defenses}

Existing defenses in FL rely on either robust aggregation functions~\cite{KRUM, Bulyan,xie2018phocas} that exclude some of the outlying client updates from the model, or gradient clipping techniques~\cite{sun2019can,gupta2021byzantine,hong2020effectiveness} that limit the contributions of each client to the model update. We select a defense from each class: Trimmed Mean~\cite{xie2018phocas} and gradient clipping~\cite{sun2019can}, adapt them to P2PFL, and evaluate them against the backdoor attacks.
\fi

\begin{figure}[!th]
    \centering   
    \vspace{-15pt}
    \subfloat[][Test Accuracy]{%
       \includegraphics[width=0.48\linewidth]{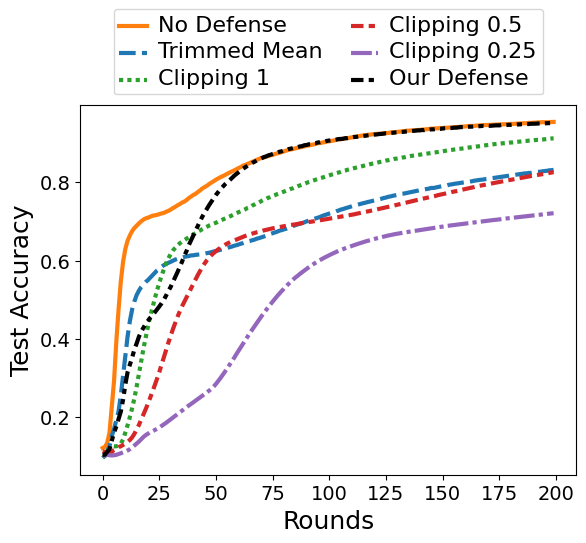}
       \label{fig:defense_acc}
       }
       \subfloat[][Attack Success]{%
    \includegraphics[width=0.48\linewidth]{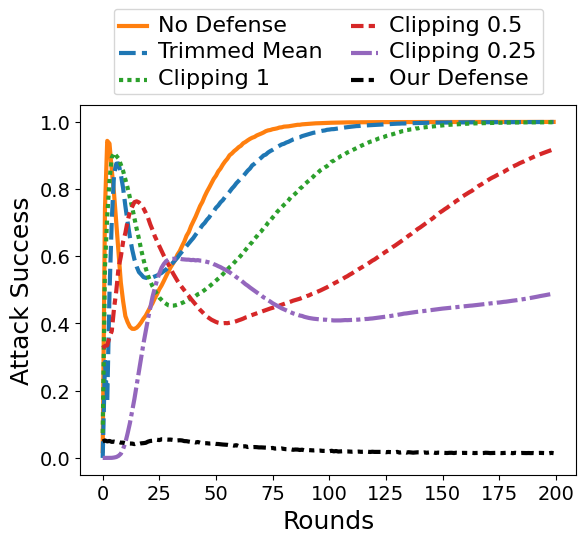}
       \label{fig:defense_attack}
       }
       \caption{Defense evaluation: (a) test accuracy and (b) attack success on poisoned samples with 60-node Watts Strogatz topology, 10\% adversarial nodes, and PageRank node selection strategy.} 
    \label{fig:defense}
    \vspace{-10pt}
\end{figure}

\iffullpaper
\else
\vspace{3pt}
\myparagraph{Robust Aggregation} 
\fi
We adapt Trimmed Mean~\cite{xie2018phocas,T-MeanT-Median_icml2018} to P2PFL, by sorting all the peer updates and filtering out  the  $p$ highest and lowest values and averaging the remaining updates.  
In our experiments (Figure~\ref{fig:defense}), we  observed for $p=1$ that Trimmed Mean is not effective against the backdoor attacks in P2PFL, as the attacker still achieves 100\% attack success. Additionally, the learning procedure is slowed down, and accuracy converges slower than the ``No Defense'' case.

\iffullpaper
\else 
\vspace{3pt}
\myparagraph{Clipping Defense}
\fi
Gradient clipping represents another standard defense against  poisoning  attacks in FL~\cite{sun2019can}. 
The most devastating poisoning attack in FL is  model poisoning~\cite{bagdasaryan2020backdoor,DBLP:conf/icml/BhagojiCMC19}, where the contribution of each compromised node is amplified by the boosting factor applied to the local model. In the extreme, a model boosting attack could overwrite the global model, and therefore gradient clipping is critical for limiting the contribution of individual clients.
In  gradient clipping in FL, the server bounds the update sent by each participant  by a threshold norm $C$ before aggregation. 
We adapt this defense to the P2PFL setting that does not rely on a trusted server to aggregate and bound updates. Instead, each peer rescales all updates which contribute to its model using:
%
\begin{equation}
    U^{t}_{j,C} = U^{t}_j/ \max(1, || U^{t}_j||/C) 
    \label{eq:bound}
\end{equation}
%

\revision{\noindent Assume $\theta^t_j$ are the model weights of peer $j$ at round $t$. $U^t_j = \theta^t_j - \theta^{t-1}_j$ is the update sent by peer $j$ at round $t$, $||U^t_j||$ is the $\ell_2$ norm of the peer update, and $C$ is the clipping norm. Selecting  the clipping norm $C$ is not straightforward, as there is a tradeoff between attack success and test accuracy. A large $C$ reduces the impact of the defense, while a small $C$ reduces the test accuracy.}
A node can generally trust its own updates, but bounding only neighbors' contributions and not its own offsets the benefit of using P2PFL instead of local training.
Similarly, setting the clipping norm too small will reduce the benefit of aggregating updates from neighbors. 
On the other hand, a large norm enables potential attacks to be aggregated into the model.


We implemented our framework in Python's deep learning API Keras using the Adam optimizer --- a stochastic gradient descent method based on adaptive estimation of first-order and second-order gradients. 
For each peer, we extract the weights of the current model, rescale them to fit within the clipped norm, and then apply the rescaled weights to update the model.
To limit the contribution of malicious peers in P2PFL we first experimented with small clipping values (0.05), but the model did not converge.
Next, we selected clipping norm values of 0.25, 0.5, and 1 and present results with these clipping norms in Figure~\ref{fig:defense}. The 0.25 norm reduces the attack success from 1 to 0.4 (Figure~\ref{fig:defense_attack}) but at a high cost on test accuracy. The larger norms impose a smaller cost on accuracy, as the attack picks up, approaching the ``No Defense'' success of 1.0.

\revision{ Another potential defense strategy would consider training models with differential privacy (DP)~\cite{DP_SGD} that involves gradient clipping and adding noise to average gradients in each batch. In theory, DP might mitigate the impact of a targeted poisoning attack that modifies a very small number of samples. In our setting,  the poisoning rate varies between 0.5\% and 50\% at each client, and because of the group privacy property differential privacy  provides vacuous guarantees in this setting. In fact, the connection between differential privacy and targeted poisoning has been made in prior work, which discussed that DP is not a full mitigation against poisoning~\cite{dp_poisoning}. In addition, it is well known that differential privacy reduces the model utility as the privacy level increases, and the tradeoff between privacy, utility, and robustness needs to be investigated in depth. 
}

\subsection{Our Defense}
\label{sec:OurDefense}



We showed in the previous section that standard gradient clipping (that  uses  a single norm for all participants) is ineffective as a defense against backdoor attacks in P2PFL settings.  
If the clipping norm for malicious peers is too large, the malicious updates will be aggregated into the local model. 
If the clipping norm for the local model is too small, the model's convergence is significantly impacted. Due to these two conflicting requirements of the clipping norm, we propose using two different clipping norm values, one for bounding the neighbor peers' updates and one for the local model.
We have the flexibility to select a smaller norm for updtaes sent by the neighboring peers and a larger norm for the local model. We choose the neighboring norm as 0.1 and the local norm as 1, after experimenting with multiple values.
In Figure~\ref{fig:defense} we study the effectiveness of using two separate clipping norms as a defense strategy against the PageRank-based poisoning attack. We observe that after 200 rounds of training, the attack success is 0, while test accuracy reaches 0.96. 
\begin{figure}[!th]
    \centering   
    \vspace{-10pt}
    \subfloat[][Test Accuracy]{%
    \includegraphics[width=0.48\linewidth]{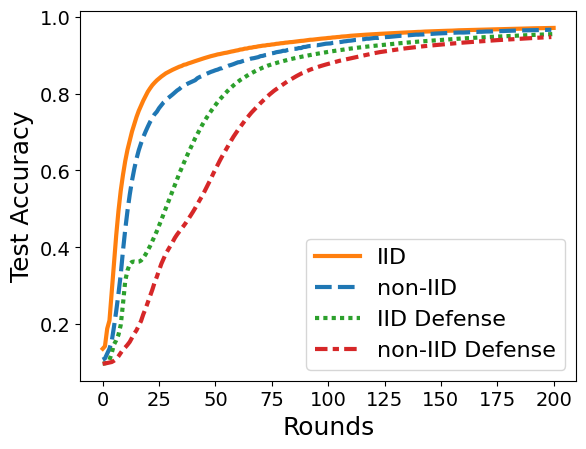}
       \label{fig:defense_noniid_acc}
       }
       \subfloat[][Attack Success]{%
    \includegraphics[width=0.48\linewidth]{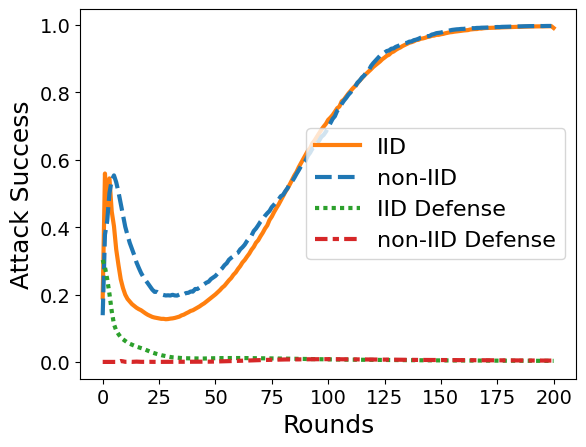}
       \label{fig:defense_noniid_attack}
       }
       \caption{Defenses in IID versus non-IID ($\alpha=1$) settings: (a) test accuracy and (b) attack success on poisoned samples. 
       Configuration: 60-node Watts Strogatz topology, 5\% adversarial nodes selected with the PageRank strategy.
       } 
    \label{fig:defense_noniid}
\end{figure}

While these experiments were carried out in IID settings, we also evaluated our defense with non-IID data distributions where the clients receive different numbers of samples per class (with $\alpha=1$). These results are represented in Figure~\ref{fig:defense_noniid}. As previously noted (Figure~\ref{fig:distrib_acc}), accuracy converges slower in non-IID even without defenses. The clipping process imposes an additional slowdown on accuracy convergence. However, our defense helps to correctly classify poisoned samples, and thus, significantly reduces the attack success in both IID and non-IID settings. Figure~\ref{fig:defense_noniid_attack} shows that within 100 rounds of training, the attack is essentially stifled (success rate of 0).

Our two-norm defense is the only strategy we are aware of that is effective against P2PFL backdoor attacks, and it obtains better accuracy under attack than Trimmed Mean and gradient clipping. The main insight behind the defense is that the trusted local model of a node is given a higher weight, while the peer models are assigned lower weight, limiting their contribution to the node's final model.
The defense can  be extended by using different weights for neighboring peers, based on the level of trust a node has for each peer.   We leave this as an exploration for future work, in addition to testing the impact of poisoning attacks and defenses in P2PFL using other datasets and model architectures. 


\revision{\myparagraph{Complex Datasets} We evaluate our defense mechanism on the CIFAR-10 dataset. In a typical centralized federated learning environment, every update from the selected clients per round contributes to the aggregation, leading to uniform changes in model parameters for all clients from round to round. In contrast, in a decentralized learning system, each peer performs its own aggregation. Consequently, the evolution of the weights for each peer varies based on factors such as the size of the local training data and the number of neighbors contributing to the aggregation. During the initial rounds, each peer's model parameters are influenced by gradients with different directions in the hyperspace. A certain number of rounds is required before the models of the peers reach a common starting point, where task accuracy begins to improve, i.e. gradients collectively have a positive effect. This phenomenon is illustrated in Figures \ref{fig:cifar10_acc} and \ref{fig:cifar10_selection_acc}. In our setup the system requires approximately 50 rounds before task accuracy starts to improve significantly. We refer to this period as the \emph{agreement} phase. During the \emph{agreement} phase, aggressive clipping can prevent the system from reaching a common starting point. Therefore, we do not use clipping during the agreement phase, and  enforce the clipping strategy after the common starting point is achieved.}

\begin{figure}[!th]
    \centering   
    \vspace{-10pt}
    \subfloat[][Test Accuracy]{%
    \includegraphics[width=0.48\linewidth]{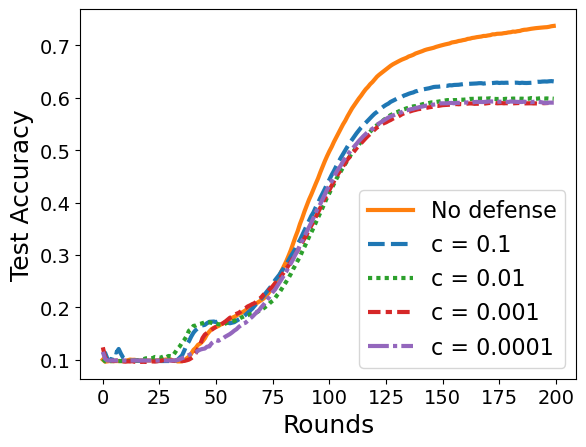}
       \label{fig:defense_iid_cifar10_acc}
       }
       \subfloat[][Attack Success]{%
    \includegraphics[width=0.48\linewidth]{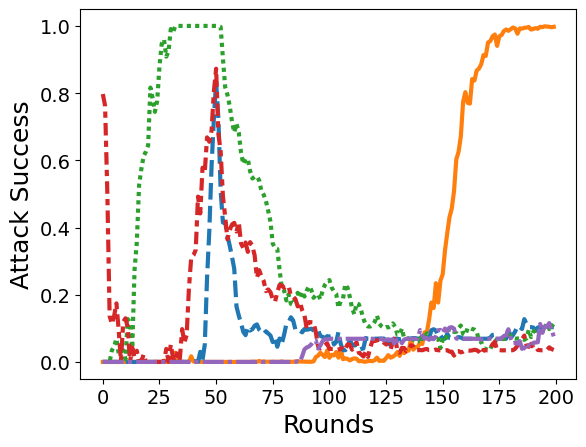}
       \label{fig:defense_iid_cifar10_attack}
       }
       \caption{\revision{Our defense with different clipping thresholds on CIFAR-10 in IID settings: (a) test accuracy and (b) attack success on poisoned samples. 
       Configuration: 30-node Watts Strogatz topology, 5\% adversarial nodes selected with the PageRank strategy.
       }} 
    \label{fig:defense_iid_cifar10}
\end{figure}
\begin{figure}[!th]
    \centering   
    \vspace{-10pt}
    \subfloat[][Test Accuracy]{%
    \includegraphics[width=0.48\linewidth]{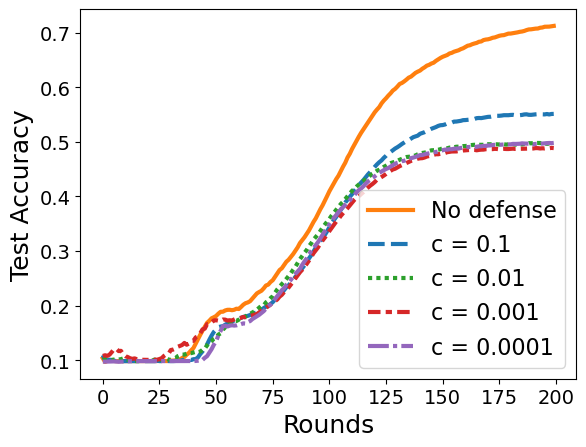}
       \label{fig:defense_noniid_cifar10_acc}
       }
       \subfloat[][Attack Success]{%
    \includegraphics[width=0.48\linewidth]{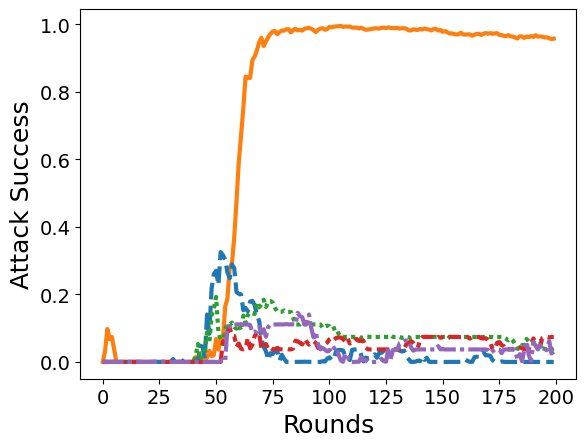}
       \label{fig:defense_noniid_cifar10_attack}
       }
       \caption{\revision{Our defense with different clipping thresholds on CIFAR-10 in non-IID settings: (a) test accuracy and (b) attack success on poisoned samples. 
       Configuration: 30-node Watts Strogatz topology, 10\% adversarial nodes selected with the PageRank strategy.
       }} 
    \label{fig:defense_noniid_cifar10}
\end{figure}

\revision{For consistency with the attack evaluation on CIFAR-10, our test setup includes 30 peers, with one peer (approximately 5\%) being adversary-controlled  and executing the backdoor attack described in this paper with a 0.005 PDR and no boosting. This scenario, featuring only one compromised peer and an extremely low PDR, represents a highly stealthy attack. We test our defense in IID and non-IID settings without local clipping and with various clipping thresholds for neighbor updates. We modify the setup for non-IID settings by increasing the number of adversary-controlled peers from 1 to 3 (5\% to 10\%) to achieve similar attack success. The results are illustrated in Figure \ref{fig:defense_iid_cifar10} and \ref{fig:defense_noniid_cifar10}. It is evident that the backdoor attack remains successful with just one adversary-controlled peer, similar to our experiments on MNIST-like datasets when no defense is deployed. However, applying our defense successfully contains the attack within 200 rounds. There is a noticeable trade-off between test accuracy and our clipping-based defense; smaller clipping thresholds negatively impact test accuracy more severely. This setup was chosen to demonstrate the efficacy of our defense against stealthy backdoor attacks such as the one demonstrated. Stronger attacks involving boosting could be mitigated with larger clipping thresholds, which would have a less significant impact on test accuracy.}

\section{Related Work}\label{sec:relwork}
We review related work in three areas: P2PFL algorithms, FL poisoning attacks, and emergent FL architectures.


\vspace{3pt}
\myparagraph{Peer-to-peer Federated Learning} 
Federated Learning\cite{mcmahan2017communication,Konecny}  trains ML models collaboratively to preserve data privacy. 
Peer-to-Peer Federated Learning is a distributed learning paradigm that removes dependence on a trusted aggregation server. 
\cite{pmlr-d2} proposed a fast algorithm for non-Byzantine settings, while several works address the Byzantine model where compromised nodes send arbitrary updates. \cite{yang2019byrdie} introduces \emph{coordinate descent} that is robust against Byzantine failures but not scalable, while \cite{fang2022bridge} designs a scalable algorithm for Byzantine faults. 
\iffullpaper
\else
\cite{pmlr-v84-bellet18a} introduces personalized Byzantine robust P2P learning for deep networks. 
\fi
\cite{he2023byzantinerobust} designs a clipping-based defense that assigns weights to neighbor contributions for aggregation and has provable convergence.
None of these works  study backdoor attacks in P2PFL.

\vspace{3pt}
\myparagraph{Model Poisoning Attacks in FL}
Poisoning attacks, including targeted and backdoor attacks have been extensively studied in traditional ML classification settings~\cite{biggio2012poisoning,mei2015using,xiao2015feature,jagielski2018manipulating, koh2017understanding, suciu2018does, shafahi2018poison,chen2017targeted,gu2019badnets}. The risk of model poisoning attacks in federated learning has been shown in several works~\cite{sun2019can, DBLP:conf/icml/BhagojiCMC19,bagdasaryan2020backdoor,wang2020attack}. In these powerful attacks, malicious clients under the adversary's control have the ability to send arbitrary model updates to the aggregation server and poison the global model.  Recent work on network-level adversaries in FL showed that adversaries might cleverly drop network packets to amplify model poisoning attacks and significantly reduce the model's performance on sub-populations~\cite{severi2022network}.

There are several defenses against model poisoning attacks proposed in centralized FL settings~\cite{FoolsGold_raid2020,FLTrust_ndss2021,Flame_usenix2022,DeepSight_ndss2022,MESAS_CCS2023}, but their applicability to P2P federated learning architectures needs to be further investigated. 

\vspace{3pt}
\myparagraph{Edge Federated Learning} 
Recently, FL has been studied in the context of edge computing~\cite{khan2020federated, lim2020federated}. Edge federated learning leverages data collected on widely dispersed edge devices, such as IoT and new 5G technologies to learn a global model shared by multiple decentralized edge clients.

\section{Broader Impact, Limitations, and Future Work}

\revision{Our work demonstrates for the first time that P2P decentralized learning systems are vulnerable to backdoor poisoning attacks. It is critical to study vulnerabilities of ML systems, including decentralized learning in P2P systems, to understand the system's attack surface and proactively investigate mitigations. This is a common cycle followed in adversarial machine learning~\cite{Adv_ML_Taxonomy}, and we initiate the study of poisoning attacks in P2PFL in this paper.  The implications of our attack in safety critical P2PFL applications could be devastating. For instance, in P2PFL for clinical applications~\cite{Swarm}, multiple hospitals engage in decentralized learning for training models that detect patients with severe  illnesses with the goal of realizing the promise of precision medicine. Our work showed that the risk of a data breach at a single hospital  could propagate  poisoning attacks at all other participating hospitals.  As more applications in P2PFL emerge, the risk of poisoning attack needs to be studied in more depth from both the learning and the application's perspective. }

\revision{We hope that our work will open up new research directions to study the integrity of ML models trained in P2PFL. We envision that more advanced attacks could emerge  with different adversarial objectives. For instance, targeting a particular client or subgroup of clients and degrading their accuracy is an interesting attack model left for future work. Additionally, there are still several challenges to address to train robust models in P2PFL architectures. We proposed an initial mitigation strategy that considers different clipping norms for the local and peer models, but the next step is to investigate an adaptive attack with knowledge of the defense and improve the defense robustness. Also, studying the design of certified defenses against poisoning such as SafeNet~\cite{SafeNet} is an interesting avenue for future work. }

\section{Conclusion}
\label{sec:conclusion}

We proposed a new modular P2PFL architecture that separates the learning and communication graphs, and allows to emulate a P2PFL system by instantiating real network topologies.  We studied  backdoors attacks in P2PFL and showed that a small number of attackers (5\%  of nodes) could achieve a high attack success, without decreasing the model's accuracy on clean data. We showed that defenses proposed for centralized FL settings, such as gradient clipping and Trimmed Mean, are ineffective in P2PFL. 
We propose a new defense that uses a weighted combination of the local model and model updates sent by a node's peer by assigning a higher weight to the trusted local model. 
Our work opens up new avenues for experimenting with other learning protocols in P2PFL architectures, and evaluating other attacks and defenses. 

\section*{Acknowledgements}

This research was supported by the Department of Defense Multidisciplinary Research Program of the University Research Initiative (MURI) under contract W911NF-21-1-0322.
\bibliographystyle{IEEEtran}
\bibliography{references}

\begin{thebibliography}{10}
\providecommand{\url}[1]{#1}
\csname url@samestyle\endcsname
\providecommand{\newblock}{\relax}
\providecommand{\bibinfo}[2]{#2}
\providecommand{\BIBentrySTDinterwordspacing}{\spaceskip=0pt\relax}
\providecommand{\BIBentryALTinterwordstretchfactor}{4}
\providecommand{\BIBentryALTinterwordspacing}{\spaceskip=\fontdimen2\font plus
\BIBentryALTinterwordstretchfactor\fontdimen3\font minus \fontdimen4\font\relax}
\providecommand{\BIBforeignlanguage}[2]{{%
\expandafter\ifx\csname l@#1\endcsname\relax
\typeout{** WARNING: IEEEtran.bst: No hyphenation pattern has been}%
\typeout{** loaded for the language `#1'. Using the pattern for}%
\typeout{** the default language instead.}%
\else
\language=\csname l@#1\endcsname
\fi
#2}}
\providecommand{\BIBdecl}{\relax}
\BIBdecl

\bibitem{car_2}
B.~R. Kiran, I.~Sobh, V.~Talpaert, P.~Mannion, A.~Sallab, S.~Yogamani, and P.~Pérez, ``Deep reinforcement learning for autonomous driving: A survey,'' \emph{IEEE Transactions on Intelligent Transportation Systems}, vol.~23, no.~6, pp. 4909--4926, 2022.

\bibitem{recommendation_1}
S.~Zhang, L.~Yao, A.~Sun, and Y.~Tay, ``Deep learning based recommender system: A survey and new perspectives,'' \emph{ACM CSUR}, 2019.

\bibitem{health_2}
G.~Manogaran and D.~Lopez, ``A survey of big data architectures and machine learning algorithms in healthcare,'' \emph{International Journal of Biomedical Engineering and Technology}, vol.~25, pp. 182--211, 2017.

\bibitem{EUdataregulations2018}
P.~Voigt and A.~Von~dem Bussche, ``The {EU} general data protection regulation ({GDPR}),'' \emph{A Practical Guide, 1st Ed., Cham: Springer}, vol.~10, no. 3152676, pp. 10--5555, 2017.

\bibitem{CCPA}
E.~L. Harding, J.~J. Vanto, R.~Clark, L.~Hannah~Ji, and S.~C. Ainsworth, ``Understanding the scope and impact of the {California Consumer Privacy Act} of 2018,'' \emph{Journal of Data Protection \& Privacy}, 2019.

\bibitem{mcmahan2017communication}
B.~McMahan, E.~Moore, D.~Ramage, S.~Hampson, and B.~A. y~Arcas, ``Communication-efficient learning of deep networks from decentralized data,'' in \emph{Artificial intelligence and statistics}.\hskip 1em plus 0.5em minus 0.4em\relax PMLR, 2017.

\bibitem{Konecny}
J.~Konečný, H.~B. McMahan, F.~X. Yu, P.~Richtarik, A.~T. Suresh, and D.~Bacon, ``Federated learning: Strategies for improving communication efficiency,'' in \emph{NIPS Workshop on Private Multi-Party ML}, 2016.

\bibitem{bonawitz2017practical}
K.~Bonawitz, V.~Ivanov, B.~Kreuter, A.~Marcedone, H.~B. McMahan, S.~Patel, D.~Ramage, A.~Segal, and K.~Seth, ``Practical secure aggregation for privacy-preserving machine learning,'' in \emph{CCS}, 2017, pp. 1175--1191.

\bibitem{brendan2018learning}
H.~B. McMahan, D.~Ramage, K.~Talwar, and L.~Zhang, ``Learning differentially private recurrent language models,'' in \emph{ICLR}, 2018.

\bibitem{NEURIPS2020_c4ede56b}
J.~Geiping, H.~Bauermeister, H.~Dr\"{o}ge, and M.~Moeller, ``Inverting gradients - how easy is it to break privacy in federated learning?'' in \emph{Advances in Neural Information Processing Systems}, 2020.

\bibitem{DBLP:conf/icml/WenGFGG22}
Y.~Wen, J.~Geiping, L.~Fowl, M.~Goldblum, and T.~Goldstein, ``Fishing for user data in large-batch federated learning via gradient magnification,'' in \emph{ICML}, vol. 162, 2022, pp. 23\,668--23\,684.

\bibitem{NIPS2017_f7552665}
\BIBentryALTinterwordspacing
X.~Lian, C.~Zhang, H.~Zhang, C.-J. Hsieh, W.~Zhang, and J.~Liu, ``Can decentralized algorithms outperform centralized algorithms? a case study for decentralized parallel stochastic gradient descent,'' in \emph{Advances in Neural Information Processing Systems}, I.~Guyon, U.~V. Luxburg, S.~Bengio, H.~Wallach, R.~Fergus, S.~Vishwanathan, and R.~Garnett, Eds., vol.~30.\hskip 1em plus 0.5em minus 0.4em\relax Curran Associates, Inc., 2017. [Online]. Available: \url{https://proceedings.neurips.cc/paper_files/paper/2017/file/f75526659f31040afeb61cb7133e4e6d-Paper.pdf}
\BIBentrySTDinterwordspacing

\bibitem{hu2019decentralized}
C.~Hu, J.~Jiang, and Z.~Wang, ``Decentralized federated learning: A segmented gossip approach,'' 2019.

\bibitem{pmlr-v84-bellet18a}
A.~Bellet, R.~Guerraoui, M.~Taziki, and M.~Tommasi, ``Personalized and private peer-to-peer machine learning,'' in \emph{AISTATS}, 2018, pp. 473--481.

\bibitem{vanhaesebrouck2017decentralized}
P.~Vanhaesebrouck, A.~Bellet, and M.~Tommasi, ``Decentralized collaborative learning of personalized models over networks,'' in \emph{Artificial Intelligence and Statistics}.\hskip 1em plus 0.5em minus 0.4em\relax PMLR, 2017, pp. 509--517.

\bibitem{lalitha2019peer}
A.~Lalitha, O.~C. Kilinc, T.~Javidi, and F.~Koushanfar, ``Peer-to-peer federated learning on graphs,'' \emph{arXiv preprint arXiv:1901.11173}, 2019.

\bibitem{fang2022bridge}
C.~Fang, Z.~Yang, and W.~U. Bajwa, ``Bridge: Byzantine-resilient decentralized gradient descent,'' \emph{IEEE T Signal Information Processing}, 2022.

\bibitem{yang2019byrdie}
Z.~Yang and W.~U. Bajwa, ``Byrdie: Byzantine-resilient distributed coordinate descent for decentralized learning,'' \emph{IEEE Trans. on Signal and Inform. Processing over Networks}, vol.~5, no.~4, pp. 611--627, 2019.

\bibitem{kuwaranancharoen2020byzantine}
K.~Kuwaranancharoen, L.~Xin, and S.~Sundaram, ``Byzantine-resilient distributed optimization of multi-dimensional functions,'' in \emph{ACC}, 2020.

\bibitem{PENG2021108020}
J.~Peng, W.~Li, and Q.~Ling, ``Byzantine-robust decentralized stochastic optimization over static and time-varying networks,'' \emph{Signal Processing}, vol. 183, p. 108020, 2021.

\bibitem{gupta2021byzantine}
N.~Gupta and N.~H. Vaidya, ``Byzantine fault-tolerance in peer-to-peer distributed gradient-descent,'' \emph{arXiv preprint arXiv:2101.12316}, 2021.

\bibitem{pmlr-d2}
H.~Tang, X.~Lian, M.~Yan, C.~Zhang, and J.~Liu, ``$d^2$: Decentralized training over decentralized data,'' in \emph{ICML}, 2018, pp. 4848--4856.

\bibitem{google_speech}
I.~McGraw, R.~Prabhavalkar, R.~Alvarez, M.~G. Arenas, K.~Rao, D.~Rybach, O.~Alsharif, H.~Sak, A.~Gruenstein, F.~Beaufays, and C.~Parada, ``Personalized speech recognition on mobile devices,'' \emph{arXiv}, 2016.

\bibitem{apple_audio}
F.~Granqvist, M.~Seigel, R.~van Dalen, A.~Cahill, S.~Shum, and M.~Paulik, ``Improving on-device speaker verification using federated learning with privacy,'' \emph{arXiv}, 2020.

\bibitem{biggio2012poisoning}
B.~Biggio, B.~Nelson, and P.~Laskov, ``Poisoning attacks against support vector machines,'' \emph{arXiv preprint arXiv:1206.6389}, 2012.

\bibitem{mei2015using}
S.~Mei and X.~Zhu, ``Using machine teaching to identify optimal training-set attacks on machine learners,'' in \emph{AAAI}, 2015.

\bibitem{xiao2015feature}
H.~Xiao, B.~Biggio, G.~Brown, G.~Fumera, C.~Eckert, and F.~Roli, ``Is feature selection secure against training data poisoning?'' in \emph{ICML}, 2015.

\bibitem{koh2017understanding}
P.~W. Koh and P.~Liang, ``Understanding black-box predictions via influence functions,'' in \emph{ICML}.\hskip 1em plus 0.5em minus 0.4em\relax PMLR, 2017, pp. 1885--1894.

\bibitem{chen2017targeted}
X.~Chen, C.~Liu, B.~Li, K.~Lu, and D.~Song, ``Targeted backdoor attacks on deep learning systems using data poisoning,'' \emph{arXiv}, 2017.

\bibitem{suciu2018does}
O.~Suciu, R.~Marginean, Y.~Kaya, H.~Daume~III, and T.~Dumitras, ``When does machine learning {FAIL}? generalized transferability for evasion and poisoning attacks,'' in \emph{USENIX Security}, 2018, pp. 1299--1316.

\bibitem{shafahi2018poison}
A.~Shafahi, W.~R. Huang, M.~Najibi, O.~Suciu, C.~Studer, T.~Dumitras, and T.~Goldstein, ``Poison frogs! targeted clean-label poisoning attacks on neural networks,'' \emph{NeurIPS}, vol.~31, 2018.

\bibitem{jagielski2018manipulating}
M.~Jagielski, A.~Oprea, B.~Biggio, C.~Liu, C.~Nita-Rotaru, and B.~Li, ``Manipulating machine learning: Poisoning attacks and countermeasures for regression learning,'' in \emph{S\&P}.\hskip 1em plus 0.5em minus 0.4em\relax IEEE, 2018, pp. 19--35.

\bibitem{gu2019badnets}
T.~Gu, K.~Liu, B.~Dolan-Gavitt, and S.~Garg, ``Badnets: Evaluating backdooring attacks on deep neural networks,'' \emph{IEEE Access}, 2019.

\bibitem{tolpegin2020data}
V.~Tolpegin, S.~Truex, M.~E. Gursoy, and L.~Liu, ``Data poisoning attacks against federated learning systems,'' in \emph{ESORICS}.\hskip 1em plus 0.5em minus 0.4em\relax Springer, 2020.

\bibitem{BhagojiCMC19}
\BIBentryALTinterwordspacing
A.~N. Bhagoji, S.~Chakraborty, P.~Mittal, and S.~B. Calo, ``Analyzing federated learning through an adversarial lens,'' in \emph{ICML}, 2019, pp. 634--643. [Online]. Available: \url{http://proceedings.mlr.press/v97/bhagoji19a.html}
\BIBentrySTDinterwordspacing

\bibitem{bagdasaryan2020backdoor}
E.~Bagdasaryan, A.~Veit, Y.~Hua, D.~Estrin, and V.~Shmatikov, ``How to backdoor federated learning,'' in \emph{AISTATS}.\hskip 1em plus 0.5em minus 0.4em\relax PMLR, 2020.

\bibitem{fang2020local}
M.~Fang, X.~Cao, J.~Jia, and N.~Gong, ``Local model poisoning attacks to {Byzantine}-robust federated learning,'' in \emph{USENIX Security}, 2020.

\bibitem{shejwalkar2021manipulating}
V.~Shejwalkar and A.~Houmansadr, ``Manipulating the byzantine: Optimizing model poisoning attacks and defenses for federated learning,'' in \emph{NDSS}, 2021.

\bibitem{sun2019can}
Z.~Sun, P.~Kairouz, A.~T. Suresh, and H.~B. McMahan, ``Can you really backdoor federated learning?'' \emph{arXiv preprint arXiv:1911.07963}, 2019.

\bibitem{wang2020attack}
H.~Wang, K.~Sreenivasan, S.~Rajput, H.~Vishwakarma, S.~Agarwal, J.-y. Sohn, K.~Lee, and D.~Papailiopoulos, ``Attack of the tails: Yes, you really can backdoor federated learning,'' \emph{NeurIPS}, vol.~33, 2020.

\bibitem{martins2010hybrid}
F.~V. Martins, E.~G. Carrano, E.~F. Wanner, R.~H. Takahashi, and G.~R. Mateus, ``A hybrid multiobjective evolutionary approach for improving the performance of wireless sensor networks,'' \emph{IEEE Sensors}, 2010.

\bibitem{jiang2019new}
Y.~Jiang, X.~Ge, Y.~Zhong, G.~Mao, and Y.~Li, ``A new small-world {IoT} routing mechanism based on {Cayley} graphs,'' \emph{IEEE Internet of Things Journal}, vol.~6, no.~6, pp. 10\,384--10\,395, 2019.

\bibitem{dong2015experimental}
Z.~Dong, Z.~Wang, W.~Xie, O.~Emelumadu, C.~Lin, and R.~Rojas-Cessa, ``An experimental study of small world network model for wireless networks,'' in \emph{IEEE Sarnoff Symposium}, 2015, pp. 70--75.

\bibitem{liu2012distributed}
C.~Liu and G.~Cao, ``Distributed critical location coverage in wireless sensor networks with lifetime constraint,'' in \emph{INFOCOM}.\hskip 1em plus 0.5em minus 0.4em\relax IEEE, 2012.

\bibitem{10.1007/3-540-45748-8_24}
J.~R. Douceur, ``The sybil attack,'' in \emph{Peer-to-Peer Systems}, P.~Druschel, F.~Kaashoek, and A.~Rowstron, Eds.\hskip 1em plus 0.5em minus 0.4em\relax Berlin, Heidelberg: Springer Berlin Heidelberg, 2002, pp. 251--260.

\bibitem{multicast_ton_2008}
A.~Walters, D.~Zage, and C.~N. Rotaru, ``A framework for mitigating attacks against measurement-based adaptation mechanisms in unstructured multicast overlay networks,'' \emph{IEEE/ACM Transactions on Networking}, vol.~16, no.~6, pp. 1434--1446, 2008.

\bibitem{pollution_streaming_2007}
\BIBentryALTinterwordspacing
P.~Dhungel, X.~Hei, K.~W. Ross, and N.~Saxena, ``The pollution attack in p2p live video streaming: measurement results and defenses,'' in \emph{Proceedings of the 2007 Workshop on Peer-to-Peer Streaming and IP-TV}, ser. P2P-TV '07.\hskip 1em plus 0.5em minus 0.4em\relax New York, NY, USA: Association for Computing Machinery, 2007, p. 323–328. [Online]. Available: \url{https://doi.org/10.1145/1326320.1326324}
\BIBentrySTDinterwordspacing

\bibitem{secure_overlay_2003}
\BIBentryALTinterwordspacing
M.~Castro, P.~Druschel, A.~Ganesh, A.~Rowstron, and D.~S. Wallach, ``Secure routing for structured peer-to-peer overlay networks,'' \emph{SIGOPS Oper. Syst. Rev.}, vol.~36, no.~SI, p. 299–314, dec 2003. [Online]. Available: \url{https://doi.org/10.1145/844128.844156}
\BIBentrySTDinterwordspacing

\bibitem{pitn_icdcs_2016}
D.~Obenshain, T.~Tantillo, A.~Babay, J.~Schultz, A.~Newell, M.~E. Hoque, Y.~Amir, and C.~Nita-Rotaru, ``Practical intrusion-tolerant networks,'' in \emph{2016 IEEE 36th International Conference on Distributed Computing Systems (ICDCS)}, 2016, pp. 45--56.

\bibitem{gossipsub_sp_2024}
A.~Kumar, M.~von Hippel, P.~Manolios, and C.~Nita-Rotaru, in \emph{IEEE Symposium on Security and Privacy}, 2024.

\bibitem{ipfs_usenix_2022}
\BIBentryALTinterwordspacing
B.~Pr{\"u}nster, A.~Marsalek, and T.~Zefferer, ``Total eclipse of the heart {\textendash} disrupting the {InterPlanetary} file system,'' in \emph{31st USENIX Security Symposium (USENIX Security 22)}.\hskip 1em plus 0.5em minus 0.4em\relax Boston, MA: USENIX Association, Aug. 2022, pp. 3735--3752. [Online]. Available: \url{https://www.usenix.org/conference/usenixsecurity22/presentation/prunster}
\BIBentrySTDinterwordspacing

\bibitem{hoffman2007survey}
K.~Hoffman, D.~Zage, and C.~Nita-Rotaru, ``A survey of attacks on reputation systems,'' 2007.

\bibitem{gu2017badnets}
T.~Gu, B.~Dolan-Gavitt, and S.~Garg, ``Badnets: Identifying vulnerabilities in the machine learning model supply chain,'' \emph{arXiv}, 2017.

\bibitem{gossipsub}
D.~Vyzovitis, Y.~Napora, D.~McCormick, D.~Dias, and Y.~Psaras, ``{GossipSub: Attack-Resilient Message Propagation in the Filecoin and ETH2.0 Networks},'' \emph{arXiv}, 2020.

\bibitem{kairouz2021advances}
P.~Kairouz, H.~B. McMahan, B.~Avent, A.~Bellet, M.~Bennis, A.~N. Bhagoji, K.~Bonawitz, Z.~Charles, G.~Cormode, R.~Cummings \emph{et~al.}, ``Advances and open problems in federated learning,'' \emph{Foundations and Trends{\textregistered} in Machine Learning}, vol.~14, no. 1--2, pp. 1--210, 2021.

\bibitem{KRUM}
P.~Blanchard, E.~M. El~Mhamdi, R.~Guerraoui, and J.~Stainer, ``Machine learning with adversaries: Byzantine tolerant gradient descent,'' \emph{Advances in Neural Information Processing Systems}, vol.~30, 2017.

\bibitem{Bulyan}
R.~Guerraoui, S.~Rouault \emph{et~al.}, ``The hidden vulnerability of distributed learning in byzantium,'' in \emph{ICML}.\hskip 1em plus 0.5em minus 0.4em\relax ACM, 2018, pp. 3521--3530.

\bibitem{xie2018phocas}
C.~Xie, O.~Koyejo, and I.~Gupta, ``Phocas: dimensional byzantine-resilient stochastic gradient descent,'' \emph{arXiv}, 2018.

\bibitem{borgatti1997structural}
S.~P. Borgatti, ``Structural holes: Unpacking {Burt}’s redundancy measures,'' \emph{Connections}, vol.~20, no.~1, pp. 35--38, 1997.

\bibitem{page1999pagerank}
L.~Page, S.~Brin, R.~Motwani, and T.~Winograd, ``The pagerank citation ranking: Bringing order to the web.'' Stanford InfoLab, Tech. Rep., 1999.

\bibitem{saramaki2007generalizations}
J.~Saram{\"a}ki, M.~Kivel{\"a}, J.-P. Onnela, K.~Kaski, and J.~Kertesz, ``Generalizations of the clustering coefficient to weighted complex networks,'' \emph{Physical Review E}, vol.~75, no.~2, p. 027105, 2007.

\bibitem{chernikova2022cyber}
A.~Chernikova, N.~Gozzi, S.~Boboila, P.~Angadi, J.~Loughner, M.~Wilden, N.~Perra, T.~Eliassi-Rad, and A.~Oprea, ``Cyber network resilience against self-propagating malware attacks,'' in \emph{ESORICS}.\hskip 1em plus 0.5em minus 0.4em\relax Springer, 2022.

\bibitem{erdos1960evolution}
P.~Erdos, A.~R{\'e}nyi \emph{et~al.}, ``On the evolution of random graphs,'' \emph{Publ. Math. Inst. Hung. Acad. Sci}, vol.~5, no.~1, pp. 17--60, 1960.

\bibitem{watts1998collective}
D.~J. Watts and S.~H. Strogatz, ``Collective dynamics of ‘small-world’networks,'' \emph{nature}, vol. 393, no. 6684, pp. 440--442, 1998.

\bibitem{albert2002statistical}
R.~Albert and A.-L. Barab{\'a}si, ``Statistical mechanics of complex networks,'' \emph{Reviews of modern physics}, vol.~74, no.~1, p.~47, 2002.

\bibitem{newman2003structure}
M.~Newman, ``The structure and function of complex networks,'' \emph{SIAM review}, vol.~45, no.~2, pp. 167--256, 2003.

\bibitem{deng2012mnist}
L.~Deng, ``The {MNIST} database of handwritten digit images for machine learning research,'' \emph{IEEE Signal Processing Magazine}, 2012.

\bibitem{xiao2017fashionmnist}
H.~Xiao, K.~Rasul, and R.~Vollgraf, ``Fashion-{MNIST}: a novel image dataset for benchmarking machine learning algorithms,'' 2017.

\bibitem{cohen2017emnist}
G.~Cohen, S.~Afshar, J.~Tapson, and A.~van Schaik, ``{EMNIST}: an extension of {MNIST} to handwritten letters,'' 2017.

\bibitem{li2022federated}
Q.~Li, Y.~Diao, Q.~Chen, and B.~He, ``Federated learning on non-iid data silos: An experimental study,'' in \emph{ICDE}.\hskip 1em plus 0.5em minus 0.4em\relax IEEE, 2022, pp. 965--978.

\bibitem{pmlr-v97-yurochkin19a}
M.~Yurochkin, M.~Agarwal, S.~Ghosh, K.~Greenewald, N.~Hoang, and Y.~Khazaeni, ``{B}ayesian nonparametric federated learning of neural networks,'' in \emph{ICML}, vol.~97.\hskip 1em plus 0.5em minus 0.4em\relax PMLR, 2019, pp. 7252--7261.

\bibitem{koloskova2020decentralized}
A.~Koloskova, T.~Lin, S.~U. Stich, and M.~Jaggi, ``Decentralized deep learning with arbitrary communication compression,'' 2020.

\bibitem{daily2018gossipgrad}
J.~Daily, A.~Vishnu, C.~Siegel, T.~Warfel, and V.~Amatya, ``Gossipgrad: Scalable deep learning using gossip communication based asynchronous gradient descent,'' 2018.

\bibitem{he2023byzantinerobust}
L.~He, S.~P. Karimireddy, and M.~Jaggi, ``Byzantine-robust decentralized learning via clippedgossip,'' 2023.

\bibitem{Wang2020FedNova}
J.~Wang, Q.~Liu, H.~Liang, G.~Joshi, and H.~V. Poor, ``Tackling the objective inconsistency problem in heterogeneous federated optimization,'' in \emph{NIPS}.\hskip 1em plus 0.5em minus 0.4em\relax ACM, 2020.

\bibitem{hong2020effectiveness}
S.~Hong, V.~Chandrasekaran, Y.~Kaya, T.~Dumitra{\c{s}}, and N.~Papernot, ``On the effectiveness of mitigating data poisoning attacks with gradient shaping,'' \emph{arXiv}, 2020.

\bibitem{T-MeanT-Median_icml2018}
\BIBentryALTinterwordspacing
D.~Yin, Y.~Chen, R.~Kannan, and P.~Bartlett, ``{B}yzantine-robust distributed learning: Towards optimal statistical rates,'' in \emph{Proceedings of the 35th International Conference on Machine Learning}, ser. Proceedings of Machine Learning Research, J.~Dy and A.~Krause, Eds., vol.~80.\hskip 1em plus 0.5em minus 0.4em\relax PMLR, 10--15 Jul 2018, pp. 5650--5659. [Online]. Available: \url{https://proceedings.mlr.press/v80/yin18a.html}
\BIBentrySTDinterwordspacing

\bibitem{DP_SGD}
\BIBentryALTinterwordspacing
M.~Abadi, A.~Chu, I.~Goodfellow, H.~B. McMahan, I.~Mironov, K.~Talwar, and L.~Zhang, ``Deep learning with differential privacy,'' in \emph{Proceedings of the 2016 ACM SIGSAC Conference on Computer and Communications Security}, ser. CCS '16.\hskip 1em plus 0.5em minus 0.4em\relax New York, NY, USA: Association for Computing Machinery, 2016, p. 308–318. [Online]. Available: \url{https://doi.org/10.1145/2976749.2978318}
\BIBentrySTDinterwordspacing

\bibitem{dp_poisoning}
M.~Jagielski and A.~Oprea, ``Does differential privacy defeat data poisoning?'' 2021.

\bibitem{severi2022network}
G.~Severi, M.~Jagielski, G.~Yar, Y.~Wang, A.~Oprea, and C.~Nita-Rotaru, ``Network-level adversaries in federated learning,'' in \emph{CNS}.\hskip 1em plus 0.5em minus 0.4em\relax IEEE, 2022.

\bibitem{FoolsGold_raid2020}
\BIBentryALTinterwordspacing
C.~Fung, C.~J.~M. Yoon, and I.~Beschastnikh, ``The limitations of federated learning in sybil settings,'' in \emph{23rd International Symposium on Research in Attacks, Intrusions and Defenses (RAID 2020)}.\hskip 1em plus 0.5em minus 0.4em\relax San Sebastian: USENIX Association, Oct. 2020, pp. 301--316. [Online]. Available: \url{https://www.usenix.org/conference/raid2020/presentation/fung}
\BIBentrySTDinterwordspacing

\bibitem{FLTrust_ndss2021}
\BIBentryALTinterwordspacing
X.~Cao, M.~Fang, J.~Liu, and N.~Z. Gong, ``Fltrust: Byzantine-robust federated learning via trust bootstrapping,'' in \emph{28th Annual Network and Distributed System Security Symposium, {NDSS} 2021, virtually, February 21-25, 2021}.\hskip 1em plus 0.5em minus 0.4em\relax The Internet Society, 2021. [Online]. Available: \url{https://www.ndss-symposium.org/ndss-paper/fltrust-byzantine-robust-federated-learning-via-trust-bootstrapping/}
\BIBentrySTDinterwordspacing

\bibitem{Flame_usenix2022}
\BIBentryALTinterwordspacing
T.~D. Nguyen, P.~Rieger, H.~Chen, H.~Yalame, H.~M{\"o}llering, H.~Fereidooni, S.~Marchal, M.~Miettinen, A.~Mirhoseini, S.~Zeitouni, F.~Koushanfar, A.~Sadeghi, and T.~Schneider, ``{FLAME}: Taming backdoors in federated learning,'' in \emph{USENIX Security Symposium}, 2022. [Online]. Available: \url{https://api.semanticscholar.org/CorpusID:263886687}
\BIBentrySTDinterwordspacing

\bibitem{DeepSight_ndss2022}
\BIBentryALTinterwordspacing
P.~Rieger, T.~D. Nguyen, M.~Miettinen, and A.~Sadeghi, ``Deepsight: Mitigating backdoor attacks in federated learning through deep model inspection,'' in \emph{29th Annual Network and Distributed System Security Symposium, {NDSS} 2022, San Diego, California, USA, April 24-28, 2022}.\hskip 1em plus 0.5em minus 0.4em\relax The Internet Society, 2022. [Online]. Available: \url{https://www.ndss-symposium.org/ndss-paper/auto-draft-205/}
\BIBentrySTDinterwordspacing

\bibitem{MESAS_CCS2023}
\BIBentryALTinterwordspacing
T.~Krau\ss{} and A.~Dmitrienko, ``{MESAS}: Poisoning defense for federated learning resilient against adaptive attackers,'' in \emph{Proceedings of the 2023 ACM SIGSAC Conference on Computer and Communications Security}, ser. CCS '23.\hskip 1em plus 0.5em minus 0.4em\relax New York, NY, USA: Association for Computing Machinery, 2023, p. 1526–1540. [Online]. Available: \url{https://doi.org/10.1145/3576915.3623212}
\BIBentrySTDinterwordspacing

\bibitem{khan2020federated}
L.~U. Khan, S.~R. Pandey, N.~H. Tran, W.~Saad, Z.~Han, M.~N. Nguyen, and C.~S. Hong, ``Federated learning for edge networks: Resource optimization and incentive mechanism,'' \emph{IEEE COMMAG}, vol.~5, 2020.

\bibitem{lim2020federated}
W.~Y.~B. Lim, N.~C. Luong, D.~T. Hoang, Y.~Jiao, Y.-C. Liang, Q.~Yang, D.~Niyato, and C.~Miao, ``Federated learning in mobile edge networks: A comprehensive survey,'' \emph{IEEE Commun. Surv. Tutor.}, 2020.

\bibitem{Adv_ML_Taxonomy}
\BIBentryALTinterwordspacing
A.~Vassilev, A.~Oprea, A.~Fordyce, and H.~Andersen, ``\BIBforeignlanguage{en}{Adversarial machine learning: A taxonomy and terminology of attacks and mitigations},'' 2024-01-04 05:01:00 2024. [Online]. Available: \url{https://tsapps.nist.gov/publication/get_pdf.cfm?pub_id=957080}
\BIBentrySTDinterwordspacing

\bibitem{Swarm}
S.~W.-H. et~al., ``Swarm learning for decentralized and confidential clinical machine learning.'' \emph{Nature}, 2021.

\bibitem{SafeNet}
H.~Chaudhari, M.~Jagielski, and A.~Oprea, ``Safenet: The unreasonable effectiveness of ensembles in private collaborative learning,'' in \emph{2023 IEEE Conference on Secure and Trustworthy Machine Learning (SaTML)}, 2023, pp. 176--196.

\end{thebibliography}

\end{document}